\documentclass[letterpaper]{article} 
\usepackage{aaai23}  
\usepackage{times}  
\usepackage{helvet}  
\usepackage{courier}  
\usepackage[hyphens]{url}  
\usepackage{graphicx} 
\usepackage{natbib}  
\usepackage{caption} 
\usepackage{algorithm}
\usepackage{algorithmic}

\usepackage{newfloat}
\usepackage{listings}
\DeclareCaptionStyle{ruled}{labelfont=normalfont,labelsep=colon,strut=off} 
\floatstyle{ruled}
\newfloat{listing}{tb}{lst}{}
\floatname{listing}{Listing}
\title{Domain Adaptation with Adversarial Training on Penultimate Activations}
\author {
	Tao Sun \textsuperscript{\rm 1},
	Cheng Lu \textsuperscript{\rm 2},
	Haibin Ling \textsuperscript{\rm 1}
}
\affiliations{
	\textsuperscript{\rm 1} Stony Brook University, USA\\
	\textsuperscript{\rm 2} XPeng Motors, USA\\
	{\{tao,hling\}@cs.stonybrook.edu, luc@xiaopeng.com}
}

\usepackage{bibentry}

\usepackage{bm}
\usepackage{array}
\usepackage{amsmath,amssymb}
\usepackage{color}
\usepackage{booktabs}
\usepackage{array}
\usepackage{multirow}
\usepackage{stmaryrd}
\usepackage{colortbl}
\usepackage{threeparttable}

\newcommand{\ngrad}[1]{\underline{#1}}
\newcommand{\HL}[1]{\textcolor[rgb]{0.8,0.31,0.2235 }{\textbf{#1}}}

\def\d{\bm{d}}
\def\r{\bm{r}}
\def\x{\bm{x}}
\def\z{\bm{z}}
\definecolor{tbgray}{gray}{.9}
\definecolor{tbgray}{rgb}{1,0.8941,0.8823}

\newcommand{\eg}[0]{\textit{e.g.}}

\newcommand{\ie}[0]{\textit{i.e.}}

\begin{document}

\maketitle

\begin{abstract}
	
Enhancing model prediction confidence on target data is an important objective in Unsupervised Domain Adaptation (UDA). In this paper, we explore adversarial training on penultimate activations, \ie, input features of the final linear classification layer. We show that this strategy is more efficient and better correlated with the objective of boosting prediction confidence than adversarial training on input images or intermediate features, as used in previous works. Furthermore, with activation normalization commonly used in domain adaptation to reduce domain gap, we derive two variants and systematically analyze the effects of normalization on our adversarial training. This is illustrated both in theory and through empirical analysis on real adaptation tasks. Extensive experiments are conducted on popular UDA benchmarks under both standard setting and source-data free setting. The results validate that our method achieves the best scores against previous arts. Code is available at \HL{\url{https://github.com/tsun/APA}}.
\end{abstract}

\section{Introduction}
Unsupervised Domain Adaptation (UDA) aims to transfer knowledge from a label rich source domain to an unlabeled target domain~\cite{pan2009survey,wang2018deep,wilson2020survey}. The two domains are relevant, yet there is often a distribution shift between them. Due to the domain gap, models from the source domain often predict incorrectly or less confidently on the target domain. Thus, how to enhance model prediction confidence on target data becomes critical in UDA.

The mainstream paradigm for UDA is \emph{feature adaptation}~\cite{sun2016deep,tzeng2014deep,ganin2015unsupervised,tzeng2017adversarial}, which learns domain-invariant features. Despite its popularity, this paradigm has some intrinsic limitations in the situation with label distribution shift~\cite{li2020rethinking,prabhu2021sentry}. Recently, \emph{self-training}~\cite{zhu2005semi} has drawn increasing attention in UDA, which applies \emph{consistency regularization} or \emph{pseudo-labeling} loss on the unlabeled target data. Self-training effectively enhances prediction confidence on target data, showing superior results~\cite{liang2020we,prabhu2021sentry,liu2021cycle,sun2022prior}.

Adversarial Training~\cite{goodfellow2014explaining,jeddi2020learn2perturb,shu2021encoding} aims to improve model's robustness by injecting adversarial examples. Originally, adversarial training is proposed for supervised learning and requires the ground-truth class labels. \citet{miyato2018virtual} generalize it to semi-supervised learning and propose Virtual Adversarial Training (VAT). VAT can be viewed as a consistency regularization, whose main idea is to force the model to make similar predictions on clean and perturbed images. VAT is later introduced into domain adaptation to improve local smoothness~\cite{shu2018dirt}. Despite its popularity~\cite{cicek2019unsupervised,kumar2018co,lee2019drop}, VAT is commonly used as an auxiliary loss. In contrast, training with VAT alone often leads to poor performance on domain adaptation benchmarks~\cite{liu2021cycle,lee2019drop,zhang2021semi}. Another drawback is that VAT requires an additional back-propagation through the entire network to obtain gradients, which increases the computation cost.

The above observations motivate us to seek more effective ways of adversarial training for UDA problems. A common practice in UDA is to split the training process into a source-training stage and a target adaptation stage. Due to the relevance between the two domains, we find that weights of the linear classification layer change quite slowly in the second stage. In fact, some source-data free works~\cite{liang2020we} freeze the classifier during adaptation and focus on improving feature extractor. Since our goal is to enhance prediction confidence on unlabeled target samples, adversarial training on \emph{penultimate activations}, \ie, the input features of the final linear classification layer, is more correlated with this goal than adversarial training on input images or intermediate features. Our analysis shows that adversarial perturbations on other layer representations can be mapped into perturbations on penultimate activations. The mapping maintains the value of adversarial loss but leads to much higher accuracies. This indicates that it is more effective to manipulate the penultimate activations. Moreover, approximating the optimal adversarial perturbation for penultimate activations is also much easier as it only involves the linear classifier in the optimization problem.

We propose a framework of domain adaptation with adversarial training on penultimate activations. The method is applicable to both standard and source-data free setting. As activation normalization with $\ell_2$ norm is commonly used in UDA to reduce domain gap~\cite{xu2019larger,prabhu2021sentry}, we derive two variants and systematically analyze the effects of normalization on our adversarial training. The shrinking effect on adversarial loss gradients is discussed. We also analyze the correlations among adversarial perturbations, activation gradients and actual activation changes after model update on real adaptation tasks. 


To summarize, our contributions include: 1) we propose a novel UDA framework of adversarial training on penultimate activations; 2) we systematically analyze its relations and advantages over previous adversarial training on input images or intermediate features; and 3) we conduct extensive experiments to validate the superior performance of our method under both standard and source-data free settings.

\section{Related Work}
\textbf{Unsupervised Domain Adaptation.} Many UDA methods have been proposed recently~\cite{pan2009survey,wang2018deep,wilson2020survey}. A mainstream paradigm is feature adaptation that learns domain invariant feature representations. Some approaches minimize domain discrepancy statistics~\cite{long2017deep,sun2016deep}, and others utilize the idea of adversarial learning by updating feature extractor to fool a domain discriminator~\cite{tzeng2014deep,ganin2015unsupervised,tzeng2017adversarial}. Despite the popularity of this paradigm, recent works~\cite{li2020rethinking,prabhu2021sentry} show its intrinsic limitations when the label distribution changes across domains. On the other hand, self-training has become a promising paradigm, which adopts the ideas from semi-supervised learning to exploit unlabeled data. One line of pseudo-labeling uses pseudo-labels generated by the model as supervision to update itself~\cite{liang2020we,liu2021cycle,sun2022prior}. Another line of consistency regularization improves local smoothness using pairs of semantically identical predictions~\cite{sun2022safe}. Self-training does not force to align source and target domains. Hence it alleviates negative transfer under a large domain shift.


\noindent\textbf{Adversarial training.} 
Adversarial training aims to produce robust models by injecting adversarial examples~\cite{goodfellow2014explaining,jeddi2020learn2perturb,shu2021encoding}. Many adversarial attack methods have been designed to generate adversarial examples. Among them, gradient-based attack is a major category that obtains adversarial perturbations from adversarial loss gradients on clean data~\cite{goodfellow2014explaining,madry2017towards}. Traditionally, the labeling information is required to generate perturbation directions. \citet{miyato2018virtual} propose Virtual Adversarial Training (VAT) for semi-supervised learning. Adversarial training can be applied on input images~\cite{goodfellow2014explaining,miyato2018virtual} or intermediate features~\cite{jeddi2020learn2perturb,shu2021encoding}. Recently, \citet{chen2022adversarial} show that adversarial augmentation on intermediate features yields good performance across diverse visual recognition tasks.

\noindent\textbf{Adversarial training in DA.} \citet{shu2018dirt} first use VAT to incorporate the locally-Lipschitz constraint in conditional entropy minimization. VAT is also used as a smooth regularizer in~\cite{cicek2019unsupervised,kumar2018co,lee2019drop,li2020model}. \citet{jiang2020bidirectional} devise a bidirectional adversarial training network for SSDA, and penalize VAT with entropy. \citet{liu2019transferable} generate transferable examples to fill in the domain gap in the feature space. \citet{lee2019drop} propose to learn discriminative features through element-wise and channel-wise adversarial dropout. \citet{kim2020attract} perturb target samples to reduce the intra-domain discrepancy in SSDA. 


\section{Approach}

\subsection{Notations and Preliminaries}
In UDA, there is a source domain $\mathcal{P}_s(\mathcal{X},\mathcal{Y})$ and a target domain $\mathcal{P}_t(\mathcal{X}, \mathcal{Y})$, where $\mathcal{X}$ and $\mathcal{Y}$ are the input image space and the label space respectively. We have access to labeled source domain samples $\mathcal{D}_s=\{(\bm{x}_i^s, y_i^s)\}_{i=1}^{n_s}$ and unlabeled target domain samples $\mathcal{D}_t=\{(\bm{x}_i^t)\}_{i=1}^{n_t}$. The goal is to learn a classifier $h=g\circ f$, where $f:\mathcal{X}\rightarrow \mathcal{Z}$ denotes the feature extractor, $g:\mathcal{Z}\rightarrow \mathcal{Y}$ denotes the class predictor, and $\mathcal{Z}$ is the latent feature space. In particular, we assume that $g$ is a linear classifier with trainable weights $W$ and biases $B$, \ie, $g(z)=Wz+B$.  $z=f(\bm{x})$ is then called the penultimate activation~\cite{seo2021distribution}.


VAT~\cite{miyato2018virtual} smooths the model in semi-supervised learning by applying adversarial training on unlabeled data. Its objective is 
\begin{equation}\label{eq:loss_i}
	\!
	\begin{gathered}
		\ell(\bm{x})=D\big[ p(y|\x), p(y|\x+\r^{(v)}) \big] \\
		\mathrm{ with~~} \r^{(v)}\triangleq\mathop{\arg\max}_{\|\r \|_2\leq \epsilon}\  \ell_r(\r) = D\big[ p(y|\bm{x}), p(y|\bm{x}+\r) \big]\!\!\!
	\end{gathered}    
\end{equation}
where $D[\cdot,\cdot]$ is Kullback-Leibler divergence. Let $\overline{\rm overlines}$ indicate $\ell_2$ normalization, \eg, $\overline{\z}=\z/\|\z\|_2$ for a vector $\z$.  In~\cite{miyato2018virtual},  $\r^{(v)}$ is approximated by
\begin{equation}
	\r^{(v)}=\epsilon \overline{\nabla_{\r} \ell_r(\r)|_{\r=\xi \bm{d}}}  \label{eq:approx.vat}
\end{equation}
in which $\epsilon$ is the perturbation magnitude, $\xi$ is a small constant and $\bm d$ is a random unit vector. The perturbation is ``virtually" adversarial because it does not come from the ground-truth labels.

\begin{figure}[!t]
	\begin{center}
		\centering
		\includegraphics[width=1.0\linewidth]{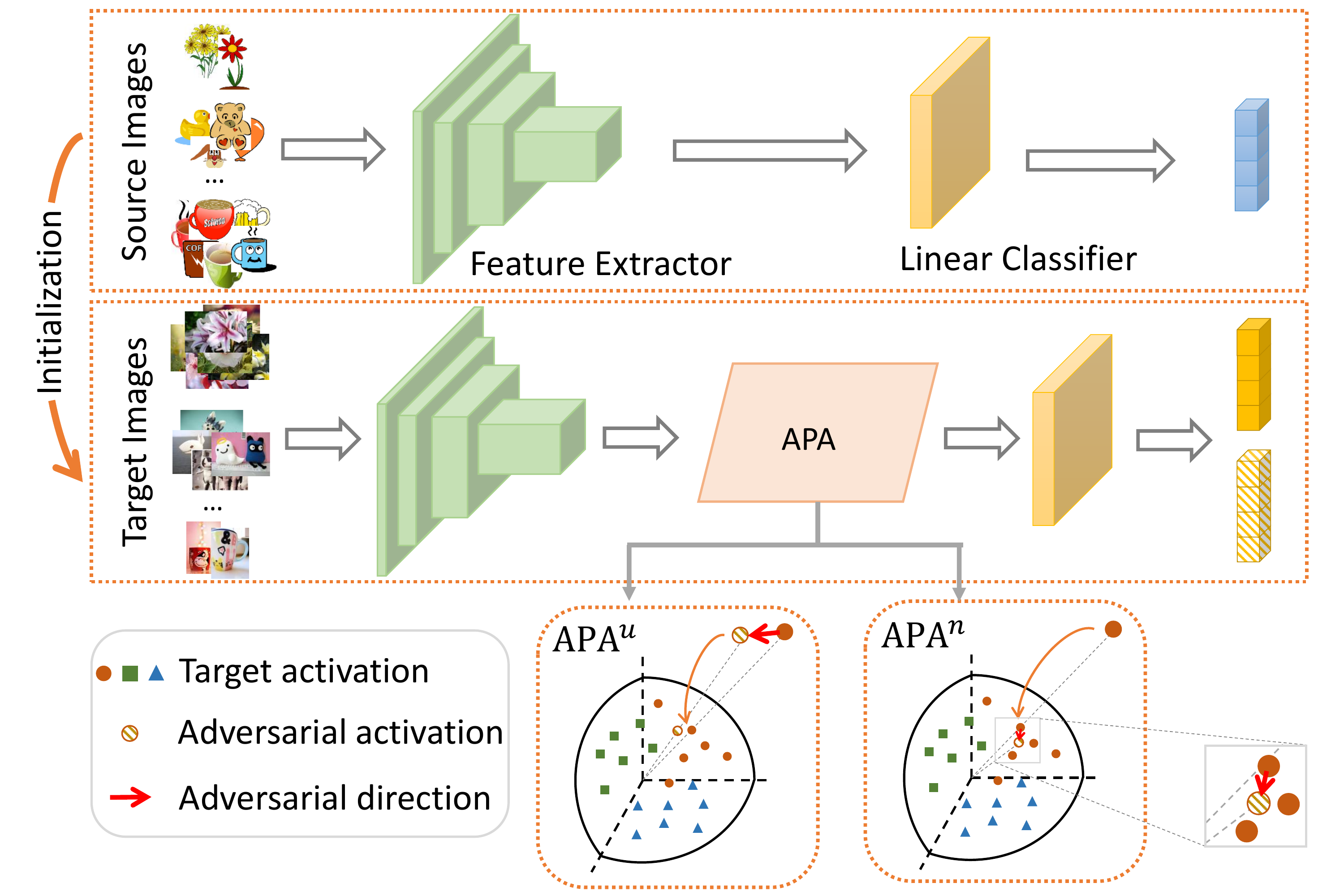}
	\end{center}
	\caption{Framework of the proposed Adversarial Training on Penultimate Activations. APA$^u$ and APA$^n$ apply adversarial perturbations on the un-normalized and normalized activations, respectively.}
	\label{fig:framework}
\end{figure}

\subsection{Adversarial Training on Penultimate Activations} 
Instead of adversarial training on input images or intermediate features, we propose Adversarial training on Penultimate Activations (APA) for unsupervised domain adaptation. We first describe motivations and details of our method, along with two variants with activation normalization in this section. Then we will discuss its relation and advantages over adversarial training on other layer representations, and make an in-depth analysis in the next section.

\noindent\paragraph{Two-stage training pipeline.}
In consistent with previous works~\cite{prabhu2021sentry,liang2020we}, we split the training process into a source-training stage and a target adaptation stage, shown in Fig.~\ref{fig:framework}. During the first stage, models are trained on source data only with standard cross entropy loss. In the second stage, the obtained source models are adapted to the target domain using target data (and source data when available). The adversarial training is applied in the second stage to enhance prediction confidence on unlabeled target samples.

\begin{figure}[!t]
	\centering
	\includegraphics[width=0.495\linewidth]{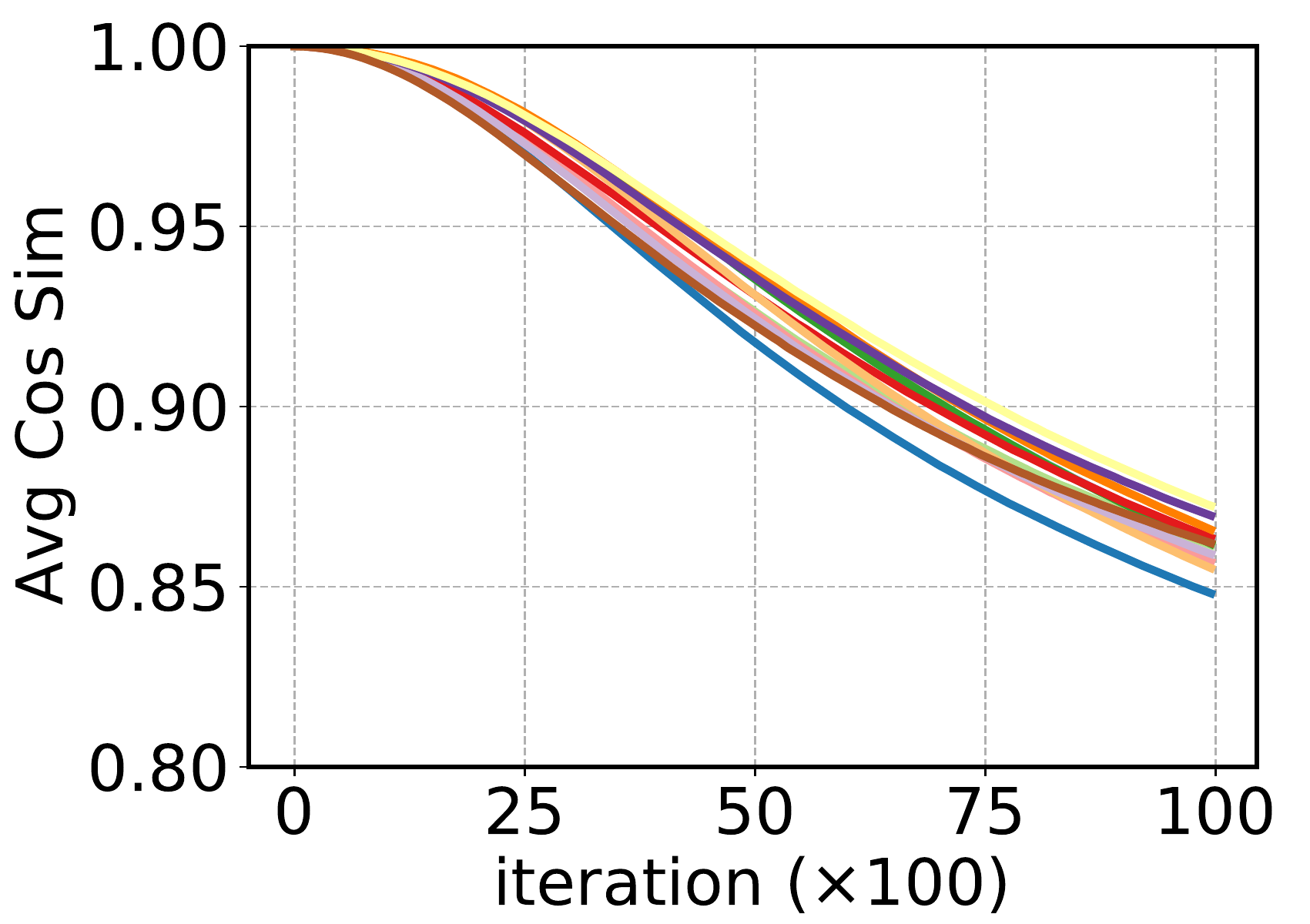}
	\includegraphics[width=0.495\linewidth]{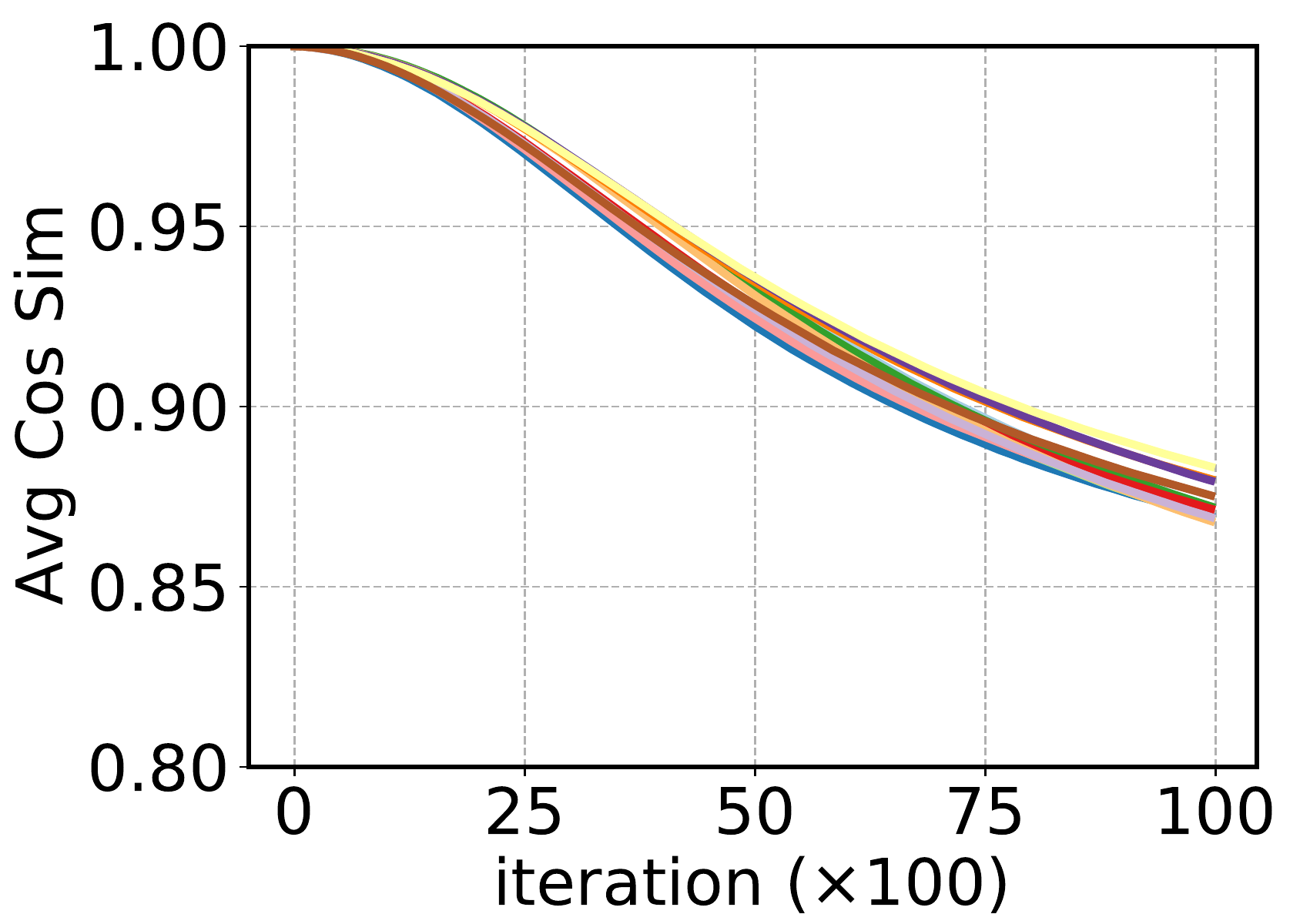}
	\caption{Averaged cosine similarity between initial classifier weights $W^{(0)}$ and current weights $W$ on 12 tasks of Office-Home under (left) standard and (right) source-data free settings. }
	\label{fig:w_corr}
\end{figure}

\noindent\paragraph{Motivations.} Since the classifier is initialized with source data and the two domains are relevant, we first want to know how fast the classifier weights change during the second stage. In Fig.~\ref{fig:w_corr}, we plot the averaged cosine similarity between initial classifier weights $W^{(0)}$ and the weights $W$ during training on 12 tasks of Office-Home. It shows that the weights change quite slowly. Therefore, it is reasonable to assume that the decision boundaries of $g$ change negligibly within a short period of training epochs. Some work~\cite{liang2020we} freeze the classifier during adaptation. We do not choose that, but find freezing the classifier does not make much difference to the performance (\textit{c.r.} Tab.~\ref{tab:freeze}). 

To improve prediction confidence on unlabeled target data, a natural way is to move their penultimate activations away from the decision boundaries. This can be realized through adversarial training. Alternatively, we can apply adversarial training on input images or intermediate features to update the penultimate activations in an indirect manner. But this would be less effective. We will discuss this in detail in the approach analysis section.

\begin{table*}[!t]
	\footnotesize
	\caption{Accuracies (\%) on \textbf{Office-Home}  (ResNet-50).}	
	\centering
	\scalebox{0.85}{
		\begin{tabular}{p{1cm}p{2cm}p{1.0cm}<{\centering}p{0.8cm}<{\centering}p{0.8cm}<{\centering}p{0.8cm}<{\centering}p{0.8cm}<{\centering}p{0.8cm}<{\centering}p{0.8cm}<{\centering}p{0.8cm}<{\centering}p{0.8cm}<{\centering}p{0.8cm}<{\centering}p{0.8cm}<{\centering}p{0.8cm}<{\centering}>{\columncolor{tbgray}}p{0.8cm}<{\centering}}
			\toprule
			& Method & A$\shortrightarrow$C & A$\shortrightarrow$P & A$\shortrightarrow$R & C$\shortrightarrow$A & C$\shortrightarrow$P & C$\shortrightarrow$R &  P$\shortrightarrow$A & P$\shortrightarrow$C & P$\shortrightarrow$R & R$\shortrightarrow$A & R$\shortrightarrow$C & R$\shortrightarrow$P & Avg.   \\ 	
			\midrule
			\multirow{12}{*}{\rotatebox{0}{UDA}} & ResNet-50 & 34.9 & 50.0 & 58.0 & 37.4 & 41.9 & 46.2 & 38.5 & 31.2 & 60.4 & 53.9 & 41.2 & 59.9 & 46.1\\
			& DANN  & 45.6 & 59.3 & 70.1 & 47.0 & 58.5 & 60.9 & 46.1 & 43.7 & 68.5 & 63.2 & 51.8 & 76.8 & 57.6 \\
			& CDAN & 50.7 & 70.6 & 76.0 & 57.6 & 70.0 & 70.0 & 57.4 & 50.9 & 77.3 & 70.9 & 56.7 & 81.6 & 65.8\\
			& SAFN & 52.0 & 71.7 & 76.3 & 64.2 & 69.9 & 71.9 & 63.7 & 51.4 & 77.1 & 70.9 & 57.1 & 81.5 & 67.3 \\
			& MDD & 54.9 & 73.7 & 77.8 & 60.0 & 71.4 & 71.8 & 61.2 & 53.6 & 78.1 & 72.5 & 60.2 & 82.3 & 68.1	\\
			& SENTRY & 61.8 &	77.4 &	80.1 &	66.3 &	71.6 &	74.7 &	66.8 &	\HL{63.0} &	80.9 &	74.0 &	\HL{66.3} &	84.1 &	72.2 \\
			& CST & 59.0 &	79.6 &	83.4 &	68.4 &	77.1 &	76.7 &	68.9 &	56.4 &	83.0 & 75.3 &	62.2 &	85.1 &	73.0 \\
			& VAT & 	49.1 &	75.0 &	78.6 &	58.4 &	71.4 &	72.4 &	57.0 &	46.4 &	78.2 &	69.4 &	54.0 &	82.8 &	66.1 \\		
			& APA$^u$ & 61.2 &	80.0 &	82.4 &	69.8 &	78.3 &	77.4 &	70.5 &	57.9 &	81.9 &	75.9 &	63.6 &	86.1 &	73.8 	\\
			& APA$^n$ & 62.0 &	81.2 &	82.6 &	71.5 &	\HL{80.6} &	79.3 &	71.9 &	60.1 &	\HL{83.4} &	\HL{76.9} &	64.2 &	86.1 &	75.0 \\
			& APA$^u$+FM & 63.1 &	80.6 &	82.6 &	\HL{71.8} &	79.7 &	79.4 &	71.4 &	61.5 &	82.7 &	75.9 &	65.5 &	86.3 &	75.0 \\	
			& APA$^n$+FM & \HL{64.0} &	\HL{81.6} &	\HL{83.7} &	70.9 &	80.3 &	\HL{80.3} &	\HL{72.8} &	62.2 &	83.0 &	76.8 &	65.5 &	\HL{86.8} &	\HL{75.7} \\	
			\midrule
			\multirow{8}{*}{\rotatebox{0}{SFDA}} & SHOT & 57.1 &	78.1 &	81.5 &	68.0 &	78.2 &	78.1 &	67.4 &	54.9 &	82.2 &	73.3 &	58.8 &	84.3 &	71.8\\
			& A$^2$Net & 58.4 & 79.0 & 82.4 & 67.5  & 79.3 & \HL{78.9} & 68.0 & 56.2 & 82.9 & 74.1 & 60.5 & 85.0 & 72.8 \\
			& NRC & 57.7 & \HL{80.3} & 82.0 & 68.1 & \HL{79.8} & 78.6 & 65.3 & 56.4 & 83.0 & 71.0 & 58.6 & \HL{85.6} & 72.2 \\ 
			& DIPE & 56.5 & 79.2 & 80.7 & 70.1 & \HL{79.8} & 78.8 & 67.9 & 55.1 & \HL{83.5} & 74.1 & 59.3 & 84.8 & 72.5 \\ 
			& APA$^u$ & 59.8 &	76.6 &	80.4 &	\HL{71.0} &	72.8 &	74.2 &	70.2 &	56.7 &	80.7 &	76.5 &	\HL{61.0} &	82.9 &	71.9\\
			& APA$^n$ & 59.3 &	75.3 &	81.0 &	68.3 &	76.3 &	76.6 &	\HL{72.8} &	57.5 &	81.8 &	76.1 &	57.9 &	84.1 &	72.2\\	
			& APA$^u$+FM & \HL{61.6} &	77.5 &	82.2 &	69.5 &	77.8 &	78.3 &	70.3 &	58.0 &	81.0 &	75.6 &	60.7 &	82.5 &	72.9 \\
			& APA$^n$+FM & 61.2 &	78.0 &	\HL{82.9} &	70.2 &	78.2 &	77.0 &	71.6 &	\HL{58.2} &	82.3 &	\HL{76.6} &	58.7 &	82.7 &	\HL{73.1} \\	
			\bottomrule
	\end{tabular} }
	\label{tab:officehome}
\end{table*}

\noindent\paragraph{Method formulation. }
We adopt adversarial training on penultimate activations to improve prediction confidence of target data during the adaptation stage. The objective is
\begin{equation}\label{eq:loss_p}
	\begin{split}
		\ell^{(p)}(\bm{x})&=D\big[  g_\sigma ({f(\bm{x})}),  g_\sigma ({f(\bm{x})+{\r^{(p)}}}) \big] \\
		\mathrm{ with~~}	\r^{(p)}&\triangleq \mathop{\arg\max}_{\|\r\|_2\leq \epsilon}\ \ell_r^{(p)}(\r) \\
		& = \mathop{\arg\max}_{\|\r\|_2\leq \epsilon}\ D\big[  g_\sigma ({f(\bm{x})}),  g_\sigma ({{f(\bm{x})}+\r})  \big]
	\end{split}    
\end{equation}
where $g_\sigma(\cdot)\triangleq\sigma(g(\cdot))$ applies the \texttt{softmax} operator $\sigma$ after the $g$. This \texttt{argmax}  optimization problem only involves $g$ as $f(\x)$ is fixed and $g$ is a \emph{linear} function, thus can be solved efficiently. In contrast, approximating $\r^{(v)}$ needs to back-propagate through the entire highly nonlinear deep neural network, which is often computationally expensive. $\r^{(p)}$ can be approximated by 
\begin{equation}\label{eq:approx}
	\r^{(p)}=\epsilon \cdot \overline{\nabla_{\r} \ell_{r}^{(p)}(\r) |_{\r=\xi \d}} 
\end{equation}

 The objective of APA is 
\begin{equation}
		\min_{f,g} \mathcal{L} = \mathbb{E}_{(\bm{x}_s, y_s)\overset{bal}\sim \mathcal{P}_s}\ell_{\mathrm{ce}}(\bm{x}_s, y_s) + \beta \mathbb{E}_{\bm{x}_t\overset{bal}\sim\mathcal{P}_t}  \ell^{(p)}(\bm{x}_t) 
	\label{eq:opt_APA}
\end{equation}
where $\ell_{ce}$ is cross entropy loss and $\beta$ is a hyper-parameter. Following previous works~\cite{jiang2020implicit,prabhu2021sentry} that tackle label shift, we use (pseudo) class-balanced sampling during training. 

The method can be easily extended to source-data free setting. Since source samples are unavailable during the adaptation stage, we use pseudo labels of confident target samples as an additional supervision term. The objective is
\begin{equation}
	\min_{f,g} \mathcal{L}_{\mathrm{sf}} = \mathbb{E}_{(\bm{x}_t, \hat{y}_t)\overset{bal}\sim \mathcal{P}_t}I_{\hat{p}_t}^\tau \ell_{\mathrm{ce}}(\bm{x}_t, \hat{y}_t) + \beta \ell^{(p)}(\bm{x}_t) 
	\label{eq:opt_APA_SF}
\end{equation}
where $\hat{y}_t$ is the pseudo label of $\bm{x}_t$. $I_{\hat{p}_t}^\tau$ is 1 if $max(\hat{p}_t)\geq \tau$ and 0 otherwise. $\tau$ is a confidence threshold.

\noindent\paragraph{Variants with activation normalization.}

Using $\ell_2$ normalization on penultimate activations is a common technique to reduce domain gap~\cite{xu2019larger,prabhu2021sentry}. It follows two variants of our method, APA$^u$ and APA$^n$,
that apply adversarial training on the un-normalized and normalized activations, respectively. Figure~\ref{fig:framework} illustrates the framework.

The adversarial loss for APA$^u$ is
\begin{equation}\label{eq:loss_pu}
	\begin{split}
		\ell^{(p_u)}(\bm{x}) &= D\big[  g_\sigma (\overline{f(\bm{x})}),  g_\sigma (\overline{f(\bm{x})+{\r^{(p_u)}}}) \big] \\
		\mathrm{ with~~}	\r^{(p_u)} & \triangleq\mathop{\arg\max}_{\|\r\|_2\leq \epsilon}D\big[  g_\sigma (\overline{f(\bm{x})}),  g_\sigma (\overline{{{f(\bm{x})}}+\r})  \big]
	\end{split}    
\end{equation}
and the adversarial loss for APA$^n$ is
\begin{equation}\label{eq:loss_pn}
		\begin{split}
			\ell^{(p_n)}(\bm{x}) &= D\big[  g_\sigma (\overline{f(\bm{x})}),  g_\sigma (\overline{f(\bm{x})}+{\r^{(p_n)}}) \big] \\
			\mathrm{ with~~}	\r^{(p_n)} & \triangleq\mathop{\arg\max}_{\|\r\|_2\leq \epsilon}D\big[  g_\sigma (\overline{f(\bm{x})}),  g_\sigma (\overline{\overline{{f(\bm{x})}}+\r})  \big]
		\end{split}    
\end{equation}
where the optimal perturbation $\r^{(p_u)}$ and $\r^{(p_n)}$ can be approximated in a similar way as Eq.~\ref{eq:approx}. $f(\bm{x})$ is also fixed in the \texttt{argmax} optimization problems. 

We apply an additional \emph{perturbation projection} step to ensure that $\overline{f(\bm{x})}+\r^{(p_n)}$ lies on the unit ball via
\begin{equation}\label{eq:rfn_proj}
	\r^{(p_n)} \leftarrow \overline{\overline{f(\bm{x})}+\r^{(p_n)}}-\overline{f(\bm{x})}
\end{equation}

\section{Approach Analysis}

\begin{figure}[!t]
	\centering
	\includegraphics[width=0.98\linewidth]{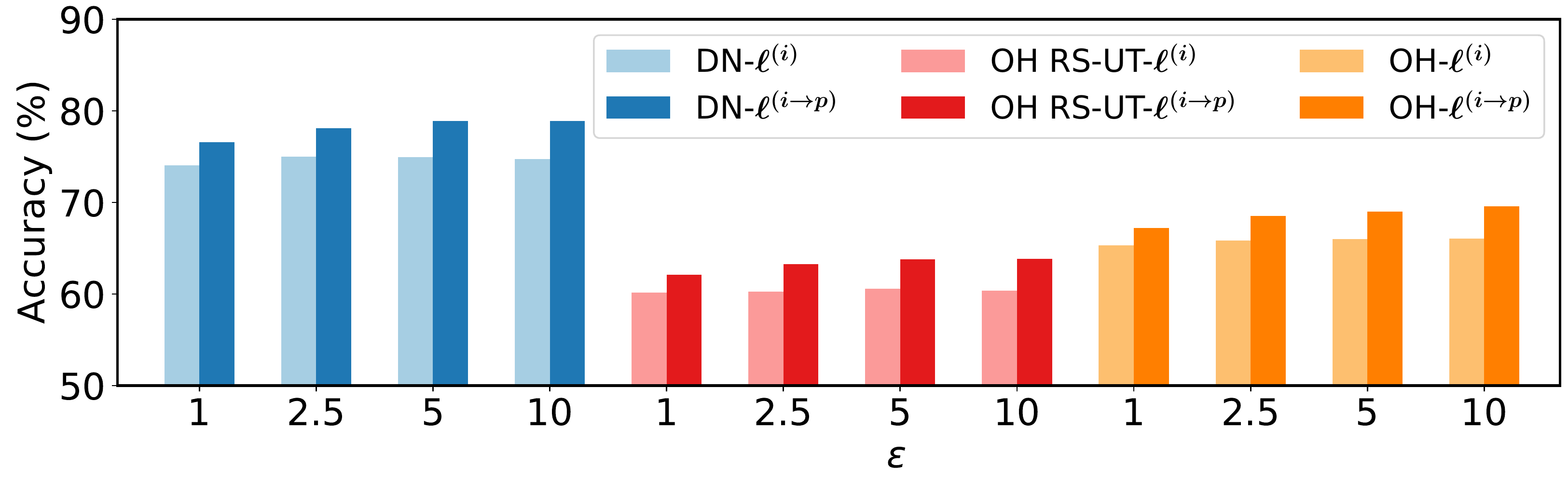}
	\includegraphics[width=0.98\linewidth]{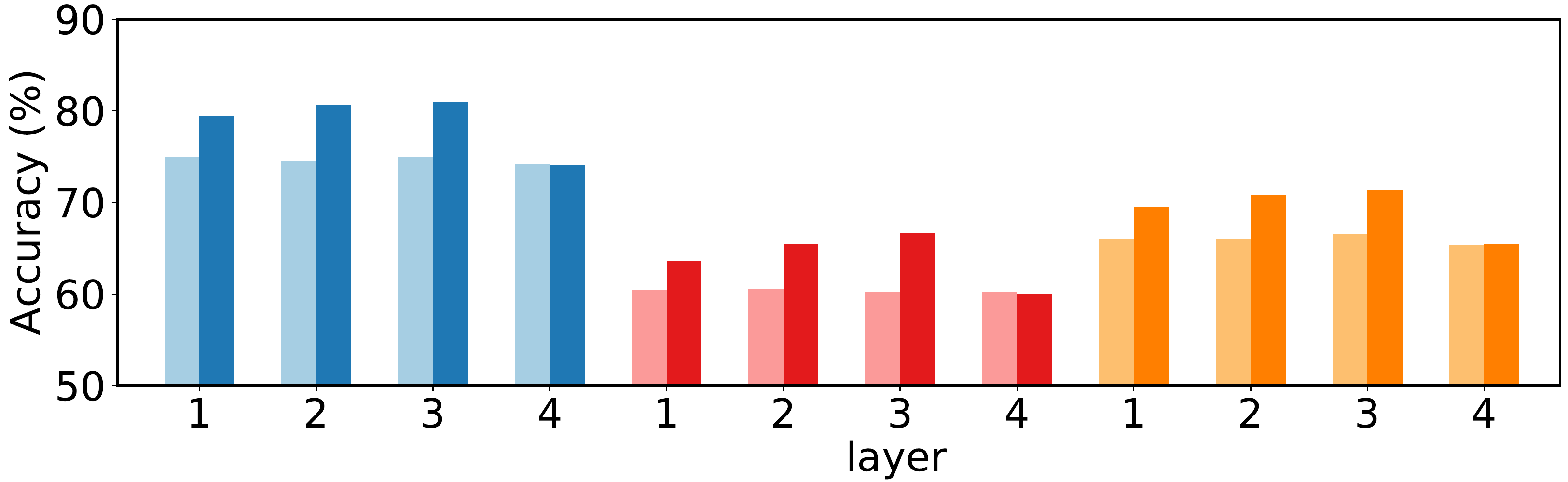}
	\caption{(Upper) Mapping input image perturbations of different magnitudes to penultimate activations. (Lower) Mapping intermediate perturbations at different layers to penultimate activations.}
	\label{fig:ptf}
\end{figure}

\begin{table*}[!t]
	\footnotesize
	\centering
	\caption{Accuracies (\%) on \textbf{VisDA}  (ResNet-101).}
	\scalebox{0.85}{
		\begin{tabular}{p{1cm}p{2cm}p{1.0cm}<{\centering}p{0.8cm}<{\centering}p{0.8cm}<{\centering}p{0.8cm}<{\centering}p{0.8cm}<{\centering}p{0.8cm}<{\centering}p{0.8cm}<{\centering}p{0.8cm}<{\centering}p{0.8cm}<{\centering}p{0.8cm}<{\centering}p{0.8cm}<{\centering}p{0.8cm}<{\centering}>{\columncolor{tbgray}}p{0.8cm}<{\centering}}
			\toprule
			& Method &  plane & bcycl & bus & car & horse & knife & mcycl & person & plant & sktbrd & train & truck & Avg. \\ 	
			\midrule
			\multirow{10}{*}{\rotatebox{0}{UDA}} & ResNet-101  & 55.1  & 53.3  & 61.9  & 59.1  & 80.6  & 17.9  & 79.7  & 31.2  & 81.0  & 26.5  & 73.5  & 8.5  & 52.4 \\
			& DANN  & 81.9  & 77.7  & 82.8  & 44.3  & 81.2  & 29.5  & 65.1  & 28.6  & 51.9  & 54.6  & 82.8  & 7.8  & 57.4 \\
			& CDAN & 85.2 & 66.9 & 83.0 & 50.8 & 84.2 & 74.9 & 88.1 & 74.5 & 83.4 & 76.0 & 81.9 & 38.0 & 73.9 \\
			& SAFN  & 93.6  & 61.3  & 84.1  & 70.6  & 94.1  & 79.0  & \HL{91.8}  & 79.6  & 89.9  & 55.6  & 89.0  & 24.4  & 76.1\\
			& SWD & 90.8 & 82.5 & 81.7 & 70.5 & 91.7 & 69.5 & 86.3 & 77.5 & 87.4 & 63.6 & 85.6 & 29.2 & 76.4 \\
			& VAT & 93.5 & 67.0 &	68.7 &	72.1 &	92.4 &	63.7 &	88.2 &	60.4 &	90.2 &	68.9 &	84.9 &	12.4 &	71.9 \\		
			& APA$^u$ & 97.1 &	90.8 &	84.5 &	72.8 &	97.4 &	96.4 &	83.9 &	83.7 &	\HL{95.8} &	87.3 &	89.8 &	42.6 &	85.2 \\
			& APA$^n$ & 96.9 &	88.3 &	83.1 &	65.9 &	95.8 &	95.3 &	85.7 &	78.8 &	94.5 &	90.1 &	88.5 &	46.4 &	84.1 \\		 
			& APA$^u$+FM & \HL{97.2} &	\HL{92.4} &	\HL{89.1} &	\HL{77.5} &	\HL{97.6} &	\HL{97.3} &	85.2 &	\HL{85.2} &	95.0 &	91.9 &	90.3 &	39.3 &	86.5 \\	
			& APA$^n$+FM & 97.0 &	90.0 &	86.4 &	76.3 &	97.2 &	\HL{97.3} &	89.4 &	82.6 &	94.5 &	\HL{94.6} &	\HL{90.8} &	\HL{52.2} &	\HL{87.4} 	\\	
			\midrule
			\multirow{8}{*}{\rotatebox{0}{SFDA}} & SHOT & 	94.3 & 88.5 & 80.1 & 57.3 & 93.1 & 94.9 & 80.7 & 80.3 & 91.5 & 89.1 & 86.3 & \HL{58.2} & 82.9 \\
			& A$^2$Net & 94.0 & 87.8 & \HL{85.6} & 66.8 & 93.7 & 95.1 & 85.8 & 81.2 & 91.6 & 88.2 & 86.5 & 56.0 & 84.3\\
			& HCL & 93.3 & 85.4 & 80.7 & \HL{68.5} & 91.0 & 88.1 & \HL{86.0} & 78.6 & 86.6 & 88.8 & 80.0 & 74.7 & 83.5 \\ 
			& DIPE & 95.2 & 87.6 & 78.8 & 55.9 & 93.9 & 95.0 & 84.1 & 81.7 & 92.1 & 88.9 & 85.4 & 58.0 & 83.1 \\
			& APA$^u$ & 96.2 &	89.9 &	83.2 &	65.7 &	95.8 &	95.3 &	81.9 &	78.2 &	91.0 &	91.5 &	88.5 &	44.1 &	83.4		\\
			& APA$^n$ & 95.2 &	88.0 &	77.3 &	64.5 &	93.2 &	96.2 &	79.2 &	77.9 &	90.7 &	89.4 &	88.1 &	45.8 &	82.1 \\	
			& APA$^u$+FM & 96.4 &	\HL{91.0} &	84.4 &	68.1 &	\HL{96.7} &	\HL{97.6} &	83.5 &	83.1 &	\HL{93.6} &	92.2 &	\HL{90.1} &	51.9 &	\HL{85.7} \\
			& APA$^n$+FM & \HL{96.5} &	90.6 &	83.4 &	64.3 &	95.8 &	97.1 &	82.3 &	\HL{83.6} &	93.3 &	\HL{93.5} &	89.6 &	55.2 &	85.4 \\
			\bottomrule
	\end{tabular} }
	\label{tab:visda}
\end{table*}

\subsection{Advantages over Adversarial Training on Input Images or Intermediate Features}

Previous works apply adversarial training on input images or intermediate features in domain adaptation. Their performances when using adversarial training alone are unsatisfactory~\cite{liu2021cycle,lee2019drop,zhang2021semi}. When the adversarial perturbation is applied far from the classifier, its impact on model prediction relies on many non-linear transformations in the remaining layers of the deep feature extractor. What is worse, the dimension of input images or intermediate features is usually large (\eg, inputs of $224\times 224\times 3$ for a typical computer vision task), making it unfavorable to obtain satisfactory approximations of optimal perturbations from the optimization view. 

We show that the perturbation on other layer representations can be mapped into perturbation on the penultimate activations. The mapping maintains the adversarial loss value but leads to higher accuracies. 

Let $f=f^b\circ f^a$ be a decomposition of the feature extractor, in which both can be an identity mapping. We use \underline{underlines} to highlight modules or variables that do not require gradient. When applying adversarial perturbation $\r^{(i)}$ on the output of $f^a$, we have
\begin{equation}
	\label{eq:vati}
	\ell^{(i)}(\bm{x}) =D\big[  g_\sigma (f(\bm{x})),\  g_\sigma (f^b(f^a(\bm{x})+\ngrad{\r^{(i)}})) \big] \\	
\end{equation}
The perturbation $\r^{(i)}$ can be mapped onto penultimate activation as $\r^{(i\rightarrow p)}\triangleq f^b(f^a(\bm{x})+{\r^{(i)}})-f(\bm{x})$, and the adversarial loss becomes
\begin{equation}\label{eq:vati2f}
	\ell^{(i\rightarrow p)}(\bm{x})= D\big[  g_\sigma (f(\bm{x})),\  g_\sigma (f(\bm{x})+\ngrad{\r^{(i\rightarrow p)}}) \big] 	
\end{equation}

Comparing Eq.~\ref{eq:vati} and Eq.~\ref{eq:vati2f}, their loss values are identical. However, the backpropagation is different due to different computation graphs. Figure~\ref{fig:ptf} shows that training with Eq.~\ref{eq:vati2f} boosts the accuracies by a large margin than Eq.~\ref{eq:vati}. The explanation is that when using $\ell^{(i\rightarrow p)}$, it directly refines the penultimate activation $f(\bm{x})$ of $\bm{x}$. In contrast, when using $\ell^{(i)}$, it is in fact updating the activation of the perturbed sample. Given that many UDA tasks work in a transductive manner where the goal is to make correct prediction for $\x$, adversarial training on representations other than penultimate activations is less correlated with this goal.


\begin{figure}[t]
	\begin{center}
		\centering
		\includegraphics[width=0.265\linewidth]{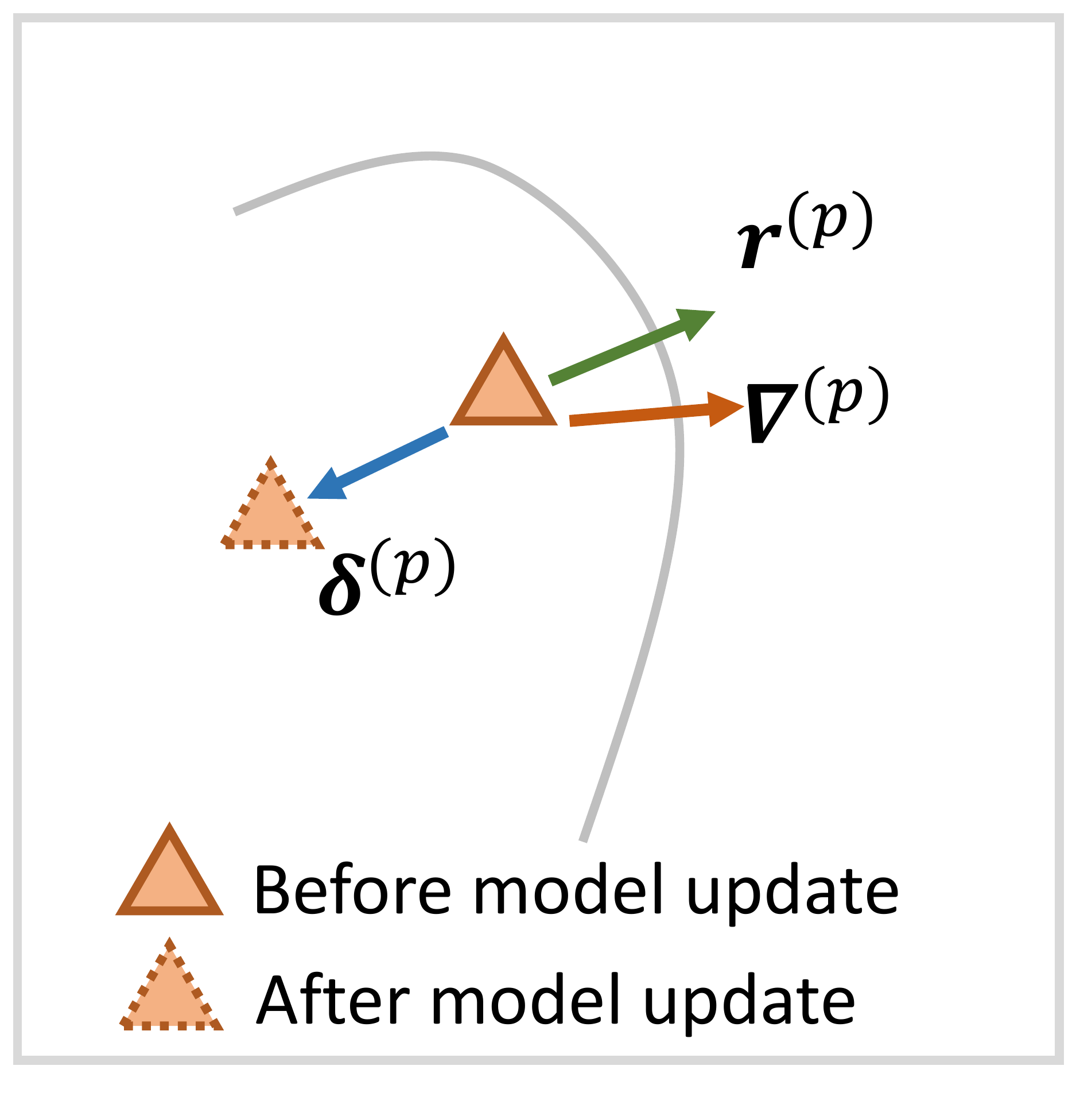}
		\includegraphics[width=0.725\linewidth]{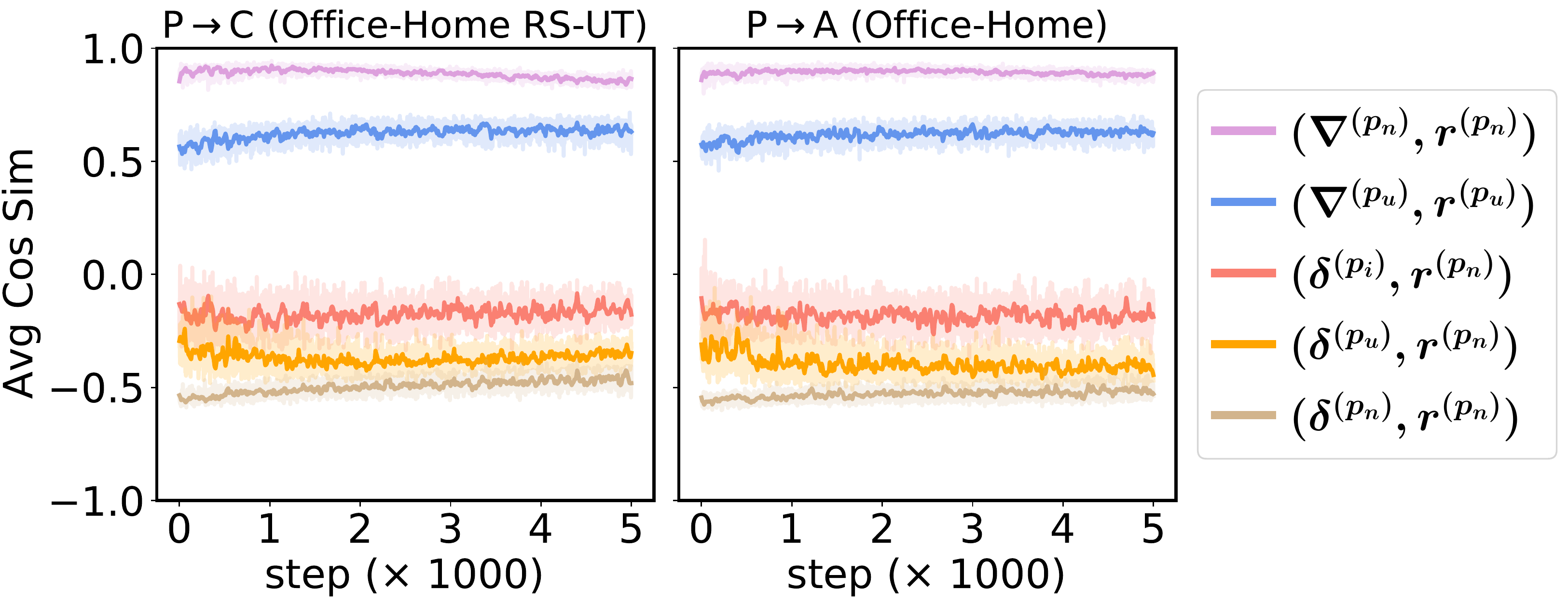}
	\end{center}
	\caption{(Left) Illustration of adversarial perturbation, activation gradient, and actual activation change. (Right) Averaged cosine similarity of these variables during the training process of two UDA tasks. (See text for details)}		
	\label{fig:corr}
\end{figure}

\subsection{Interpretation of APA} 
APA applies adversarial training on penultimate activations to improve prediction confidence. Intuitively, $f(\bm{x})$ is expected to be moved away from the boundaries. Figure~\ref{fig:corr} illustrates this idea. $\r^{(p)}$ is the adversarial perturbation, $\bm{\nabla}^{(p)}$ is the gradient of adversarial loss with respect to the penultimate activation, and $\bm{\delta}^{(p)}$ is the direction where the activation actually moves after network parameters are updated.

The two APA variants are compared with adversarial training on input images. In the right of Fig.~\ref{fig:corr}, we show the correlation among these variables during the real training processes of two UDA tasks. In the figure, $\bm{\nabla}^{(p_u)}=\partial \ell^{(p_u)}(\bm{x})/ \partial f(\bm{x})$, $\bm{\nabla}^{(p_n)}=\partial \ell^{(p_n)}(\bm{x})/ \partial \overline{f(\bm{x})}$,  $\bm{\delta}^{(p_*)}=\overline{f^*(\bm{x})}-\overline{f(\bm{x})}$, where $f^*(\bm{x})$ are the new activations after one-step gradient descend with learning rate 0.001 based on the corresponding adversarial losses. As can be seen, there is a strong positive correlation between adversarial perturbation and active gradient. In particular, $\bm{\nabla}^{(p_n)}$ is highly close to $\r^{(p_n)}$ in direction. $\bm{\delta}^{(p_*)}$ are negatively correlated with $\r^{(p_n)}$. Recall that $\r^{(p_n)}$ is the optimal adversarial direction by construction, hence we can claim that the adversarial training indeed pushes target samples away from decision boundaries. Comparing the three variables, $\bm{\delta}^{(p_n)}$ has the strongest correlation with $\r^{(p_n)}$, followed by $\bm{\delta}^{(p_u)}$ and $\bm{\delta}^{(p_i)}$. This validates that APA is more effective in improving model prediction confidence than image-based training.

\subsection{Shrinking Effect of Normalization}

APA$^u$ and APA$^n$ have different properties due to different placing of normalization. With a similar mapping trick, we can create adversarial losses with identical loss values but different gradients. Let $\r^{(p_{n\rightarrow u})} \triangleq \r^{(p_n)}\cdot \| f(\bm{x})\|$ and $\r^{(p_{u\rightarrow n})} \triangleq  \overline{f(\bm{x})+\r^{(p_u)}}-\overline{f(\bm{x})}$, it is easy to prove that $\ell^{(p_n)}|_{\r^{(p_{u\rightarrow n})}}$ equals $\ell^{(p_u)}$ and  $\ell^{(p_u)}|_{\r^{(p_{n\rightarrow  u})}}$ equals $\ell^{(p_n)}$ in terms of objective value.

\begin{figure}[!t]
	\centering
	\includegraphics[width=\linewidth]{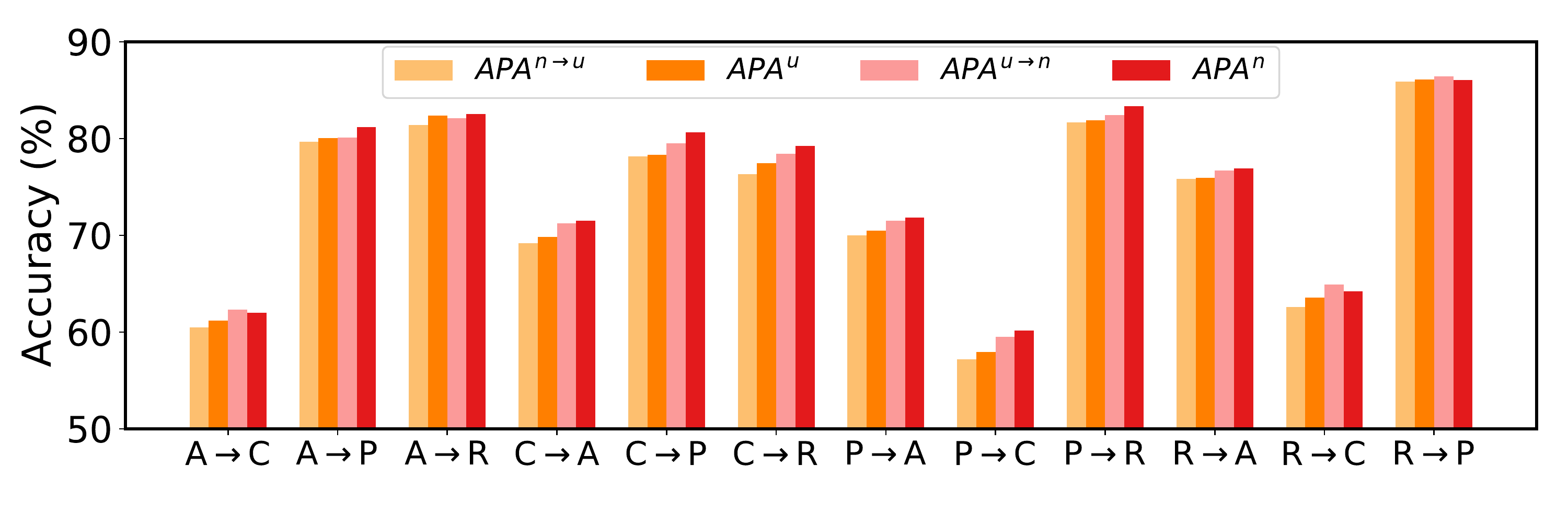}
	\caption{Comparing APA variants on Office-Home.}
	\label{fig:f2n}
\end{figure}

\begin{figure}[!t]
	\centering
	\includegraphics[width=\linewidth]{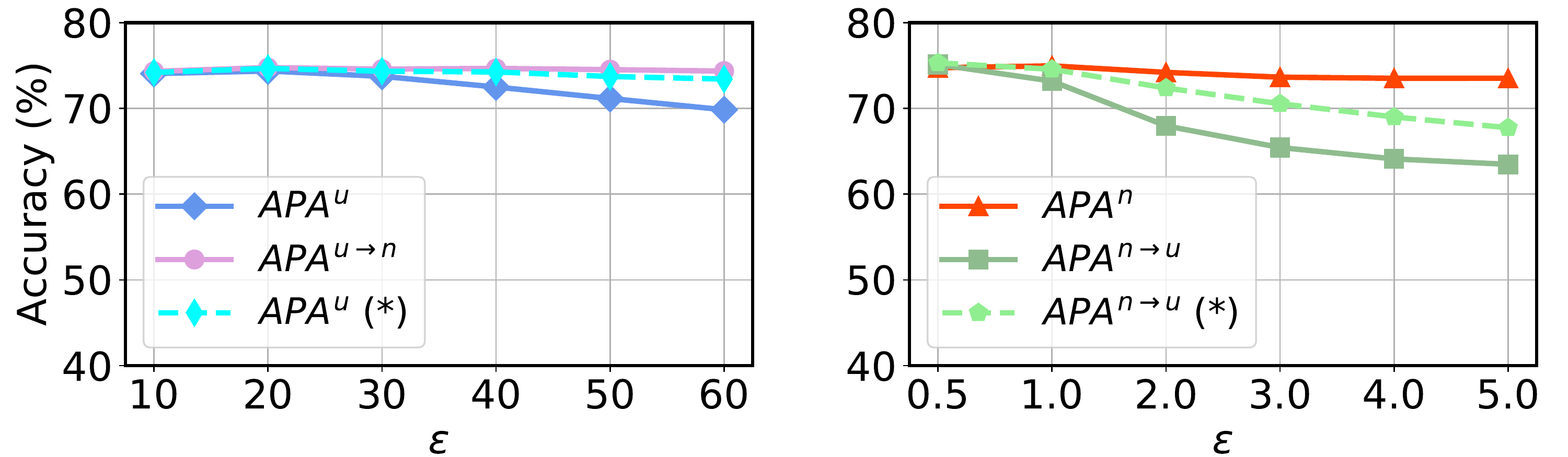}
	\caption{Effects of perturbation magnitude $\epsilon$ on Office-Home. (*) Compensated with activation norm ratio.}
	\label{fig:eps_cmp}
\end{figure}

\begin{table*}[t]
	\footnotesize
	\centering
	\caption{Per-class average accuracies (\%) on \textbf{Office-Home (RS-UT)}  (ResNet-50).}
	\scalebox{0.85}{
		\begin{tabular}{p{1cm}p{2cm}p{1.2cm}<{\centering}p{1.2cm}<{\centering}p{1.2cm}<{\centering}p{1.2cm}<{\centering}p{1.2cm}<{\centering}p{1.2cm}<{\centering}>{\columncolor{tbgray}}p{1.2cm}<{\centering}}
			\toprule
			& Method & R$\shortrightarrow$P &	R$\shortrightarrow$C &	P$\shortrightarrow$R &	P$\shortrightarrow$C &	C$\shortrightarrow$R & C$\shortrightarrow$P & Avg.   \\ 
			\midrule
			\multirow{10}{*}{\rotatebox{0}{UDA}} & ResNet-50 & 69.8 & 38.3  & 67.3  & 35.8 & 53.3 & 52.3   & 52.8 \\ 
			& DANN & 71.6 & 46.5 & 68.4 & 38.8 & 58.8 & 58.0 & 56.9 \\			
			& COAL & 73.6 & 42.6 & 73.3 & 40.6 & 59.2 & 57.3 & 58.4 \\			
			& InstaPBM & 75.6 & 42.9 & 70.3 & 39.3 & 61.8 & 63.4 & 58.9 \\
			& MDD+IA & 76.1   & 50.0  & 74.2 & 45.4  & 61.1  & 63.1  & 61.7 \\ 
			& SENTRY & 76.1 &	56.8 &	73.6 &	54.7 &	\HL{65.9} &	64.3 &	65.3 \\	
			& CST  & 79.6 &	55.6 &	\HL{77.4} &	47.1 &	61.6 &	62.0 &	63.9\\
			& VAT & 78.0 &	45.4 &	73.0 &	43.2 &	60.6 &	62.1 &	60.4 \\		
			& APA$^u$ & 80.2 &	\HL{60.3} &	76.1 &	53.8 &	65.7 &	67.0 &	\HL{67.2} \\
			& APA$^n$ & \HL{80.3} &	57.7 &	76.7 &	\HL{54.9} &	64.1 &	\HL{67.5} &	66.9 \\		
			\midrule
			\multirow{3}{*}{\rotatebox{0}{SFDA}} & SHOT & 77.0 &	50.3 &	75.9 &	47.0 &	64.3 &	64.6 &	63.2	 \\
			& APA$^u$ & \HL{77.8} &	\HL{55.8} &	\HL{76.3} &	49.7 &	\HL{65.0} &	\HL{64.7} &	\HL{64.9} \\
			& APA$^n$ & 77.3 &	\HL{55.8} &	76.2 &	\HL{50.8} &	64.5 &	64.2 &	64.8 \\
			\bottomrule
	\end{tabular} }
	\label{tab:officehome-rsut}
\end{table*}

\begin{table*}[!t]
	\footnotesize
	\caption{Per-class average accuracies (\%) on \textbf{Domainnet}  (ResNet-50).}	
	\centering
	\scalebox{0.85}{
		\begin{tabular}{p{1cm}p{2cm}p{1.0cm}<{\centering}p{0.8cm}<{\centering}p{0.8cm}<{\centering}p{0.8cm}<{\centering}p{0.8cm}<{\centering}p{0.8cm}<{\centering}p{0.8cm}<{\centering}p{0.8cm}<{\centering}p{0.8cm}<{\centering}p{0.8cm}<{\centering}p{0.8cm}<{\centering}p{0.8cm}<{\centering}>{\columncolor{tbgray}}p{0.8cm}<{\centering}}
			\toprule
			& Method & R$\shortrightarrow$C &	R$\shortrightarrow$P &	R$\shortrightarrow$S &	C$\shortrightarrow$R &	C$\shortrightarrow$P &	C$\shortrightarrow$S &	P$\shortrightarrow$R &	P$\shortrightarrow$C &	P$\shortrightarrow$S &	S$\shortrightarrow$R &	S$\shortrightarrow$C &	S$\shortrightarrow$P	 & Avg.   \\ 	
			\midrule
			\multirow{9}{*}{\rotatebox{0}{UDA}} & ResNet-50 & 58.8 &67.9 &53.1 &76.7 &53.5 &53.1 &84.4 &55.5 & 60.2 &74.6 &54.6 &57.8 &62.5\\
			& DANN &63.4 &73.6 &72.6 &86.5 &65.7 &70.6 &86.9 &73.2 &70.1 &85.7 &75.2 &70.0 &74.5\\
			& COAL &73.8 &75.4 &70.5 &89.6 &69.9 &71.3 &89.8 &68.0 &70.5 &88.0 &73.2 &70.5 &75.9\\
			& InstaPBM & 80.1 & 75.9 & 70.8 & 89.7 & 70.2 & 72.8 & 89.6 & 74.4 & 72.2 & 87.0 & 79.7 & 71.7 & 77.8 \\
			& SENTRY & 83.9 &	76.7 &	74.4 &	90.6 &	76.0 &	79.5 &	90.3 &	82.9 &	75.6 &	90.4 &	82.4 &	74.0 &	81.4 \\
			& CST & 83.9 &	78.1 &	77.5 &	90.9 &	76.4 &	79.7 &	90.8 &	82.5 &	76.5 &	90.0 &	82.8 &	74.4 &	82.0 \\
			& VAT & 73.6 &	74.8 &	65.2 &	88.1 &	66.1 &	64.5 &	88.2 &	71.4 &	70.7 &	87.3 &	77.4 &	69.9 &	74.8 \\		
			& APA$^u$ & 85.0 &	79.7 &	79.0 &	91.2 &	\HL{77.9} &	79.3 &	91.3 &	82.4 &	78.5 &	90.7 &	84.3 &	77.7 &	83.1 \\
			& APA$^n$ & \HL{85.3} &	\HL{80.4} &	\HL{80.2} &	\HL{91.8} &	77.2 &	\HL{79.9} &	\HL{91.7} &	\HL{83.7} &	\HL{80.4} &	\HL{91.2} &	\HL{85.4} &	\HL{79.3} &	\HL{83.9} \\
			\midrule
			\multirow{3}{*}{\rotatebox{0}{SFDA}} & SHOT & 79.4	 & 75.4 &	72.8 &	88.4 &	74.0 &	75.5 &	89.8 &	77.7 &	76.2 &	88.3 &	80.5 &	70.8 &	79.1  \\
			& APA$^u$ & 84.5 &	\HL{79.0} &	77.3 &	\HL{90.2} &	\HL{78.5} &	\HL{78.9} &	\HL{91.9} &	\HL{84.9} &	\HL{77.4} &	91.6 &	\HL{84.7} &	\HL{78.7} &	\HL{83.1} \\
			& APA$^n$ & \HL{85.4} &	78.8 &	\HL{78.0} &	90.1 &	78.1 &	78.4 &	91.3 &	84.8 &	76.5 &	\HL{92.8} &	83.6 &	76.4 &	82.8 \\
			\bottomrule
	\end{tabular} }
	\label{tab:domainnet}
\end{table*}

Let $\zeta(\bm{v})=\bm{v}/\|\bm{v}\|_2$ be the $\ell_2$ normalization function, $\x$ omitted, $\zeta_u\triangleq \zeta(f+\r^{(p_u)})$, and $\zeta_n\triangleq \zeta(f)+\r^{(p_n)}$, then with some derivations
\begin{eqnarray}
	&\left(\frac{\partial \ell^{(p_u)}}{\partial f}\right)^{\top} =  \left(\frac{\partial \ell^{(p_u)}}{\partial \zeta_u} \right)^{\top} \bm{J}_{\zeta}(f+\r^{(p_u)}) \\
	&\left(\frac{\partial \ell^{(p_n)}}{\partial f}\right)^{\top} = \left(\frac{\partial \ell^{(p_n)}}{\partial \zeta_n} \right)^{\top} \bm{J}_{\zeta}(f)
\end{eqnarray}
The Jacobian matrix of $\zeta(\bm{v})$ is $\bm{J}_{\zeta}(\bm{v})=(I-\bm{v}\bm{v}^{\top}/(\bm{v}^{\top} \bm{v}))/\|\bm{v}\|$. Using the mapping trick, we can have $\zeta_n=\zeta_u$ and $\partial \ell^{(p_n)}/\partial \zeta_n=\partial \ell^{(p_u)}/\partial \zeta_u$. Suppose the second term of $\bm{J}_{\zeta}$ is negligible, the gradient of $f$ is shrunk by $\|f+\r^{(p_u)} \|$ in APA$^{u}$, while intact in APA$^{n}$. Figure~\ref{fig:f2n} compares different APA variants on Office-Home, where APA$^{n\rightarrow u}$ means using perturbation $\r^{(p_{n\rightarrow u})}$ in APA$^{u}$. The same is with APA$^{u\rightarrow n}$. As can be seen, perturbing normalized activations performs slightly better. Figure~\ref{fig:eps_cmp} shows as $\epsilon$ increases (consequently $\|f+\r^{(p_u)}\|$ increases), the accuracy of perturbing un-normalized activations drops due to shrunk gradients. This can be compensated by magnifying adversarial loss with activation norm ratio of $\|f+\r^{(p_u)} \| / \|f\|$.

\section{Experiments}
\subsection{Setup}
\textbf{Datasets.} 
\textbf{Office-Home} (OH) has 65 classes from four domains: Artistic (A), Clip Art (C), Product (P), and Real-world (R). We use both the original version and the RS-UT (Reverse-unbalanced Source and Unbalanced Target) version~\cite{tan2019generalized} that is manually created to have a large label shift. \textbf{VisDA-2017}~\cite{peng2017visda} is a synthetic-to-real dataset of 12 objects. \textbf{DomainNet}~\cite{peng2019moment} (DN) is a large UDA benchmark. We use the 40-class version~\cite{tan2019generalized} from four domains: Clipart (C), Painting (P), Real (R), Sketch (S).

\noindent\textbf{Baseline methods.} We compare our proposed APA with several lines of methods. Methods for vanilla UDA include DANN~\cite{ganin2015unsupervised}, CDAN~\cite{long2018conditional}, MDD~\cite{zhang2019bridging}, SWD~\cite{lee2019sliced}, SAFN~\cite{xu2019larger}, CST~\cite{liu2021cycle}. Methods that handle label shift include COAL~\cite{tan2019generalized}, InstaPBM~\cite{li2020rethinking}, MDD+IA~\cite{jiang2020implicit}, SENTRY~\cite{prabhu2021sentry}. SHOT~\cite{liang2020we}, A$^2$Net~\cite{xia2021adaptive}, NRC~\cite{yang2021exploiting}, DIPE~\cite{wang2022exploring}, HCL~\cite{huang2021model} are methods for source-data free setting. We also compare with self-training based methods using Entropy Minimization, Mutual Information Maximization (MI)~\cite{gomes2010discriminative,shi2012information}, VAT~\cite{miyato2018virtual}, and FixMatch (FM)~\cite{sohn2020fixmatch} loss. The results summarized in Tab. \ref{tab:officehome}--\ref{tab:domainnet} are taken from the corresponding papers whenever available.

\begin{table*}[!t]
	\footnotesize
	\caption{Effects of using perturbation projection in APA$^n$ on {DomainNet}.}	
	\centering
	\scalebox{0.85}{
		\begin{tabular}{p{1.8cm}p{1.0cm}<{\centering}p{1.0cm}<{\centering}p{1.0cm}<{\centering}p{1.0cm}<{\centering}p{1.0cm}<{\centering}p{1.0cm}<{\centering}p{1.0cm}<{\centering}p{1.0cm}<{\centering}p{1.0cm}<{\centering}p{1.0cm}<{\centering}p{1.0cm}<{\centering}p{1.0cm}<{\centering}>{\columncolor{tbgray}}p{1.0cm}<{\centering}}
			\toprule
			Projection & R$\shortrightarrow$C &	R$\shortrightarrow$P &	R$\shortrightarrow$S &	C$\shortrightarrow$R &	C$\shortrightarrow$P &	C$\shortrightarrow$S &	P$\shortrightarrow$R &	P$\shortrightarrow$C &	P$\shortrightarrow$S &	S$\shortrightarrow$R &	S$\shortrightarrow$C &	S$\shortrightarrow$P	 & Avg.   \\ 
			\midrule
			$\checkmark$ & \HL{85.3} &	\HL{80.4} &	\HL{80.2} &	\HL{91.8} &	\HL{77.2} &	\HL{79.9} &	\HL{91.7} &	\HL{83.7} &	\HL{80.4} &	\HL{91.2} &	\HL{85.4} &	\HL{79.3} &	\HL{83.9} \\
			$\times$ & 84.1 &	79.5 &	79.8 &	91.3 &	76.0 &	\HL{79.9} &	90.7 &	82.0 &	79.5 &	90.8 &	83.6 &	78.7 &	83.0 \\		
			\bottomrule
	\end{tabular} }
	\label{tab:domainnet_proj}
\end{table*}

\noindent\textbf{Implementation details.}
We implement our methods with PyTorch. The pre-trained ResNet-50 or ResNet-101~\cite{he2016deep} models are used as the backbone network of $f$. Then $f$ contains a bottleneck layer with Batch Normalization. The classification head $g$ is a single Fully-Connected layer. For all tasks, we use batch size 16, $\beta=0.1$, $\tau=0.75$, $\epsilon=30$ for APA$^u$, $\epsilon=1.0$ for APA$^n$, with an only exception on VisDA where we use $\beta=0.04$ for APA$^n$ instead. More details can be found in the supp. material. 

\subsection{Results}

\textbf{Results on standard benchmarks.}
Tables~\ref{tab:officehome},\ref{tab:visda} present evaluation results on three standard benchmarks. Our method achieves consistently the best scores against previous arts under vanilla setting. The performance is further boosted when combined with FixMatch~\cite{sohn2020fixmatch}. For source-data free setting, A$^2$Net~\cite{xia2021adaptive} is a strong comparison method employing adversarial inference, contrastive matching, and self-supervised rotation. APA$^u$+FM improves +1.4\% over it on VisDA.  

\noindent\textbf{Results on label-shifted benchmarks.}
Tables~\ref{tab:officehome-rsut},\ref{tab:domainnet} present evaluation results on two benchmarks with a large label distribution shift. Our methods improve over others tailored for this setting by a large margin. Note that we adopt the same hyper-parameters for all tasks, and the performance could likely be further improved by better tuning their values.

\subsection{Analysis}\label{sec:analysis}
\begin{figure}[h]
	\begin{center}
		\centering
		\includegraphics[width=\linewidth]{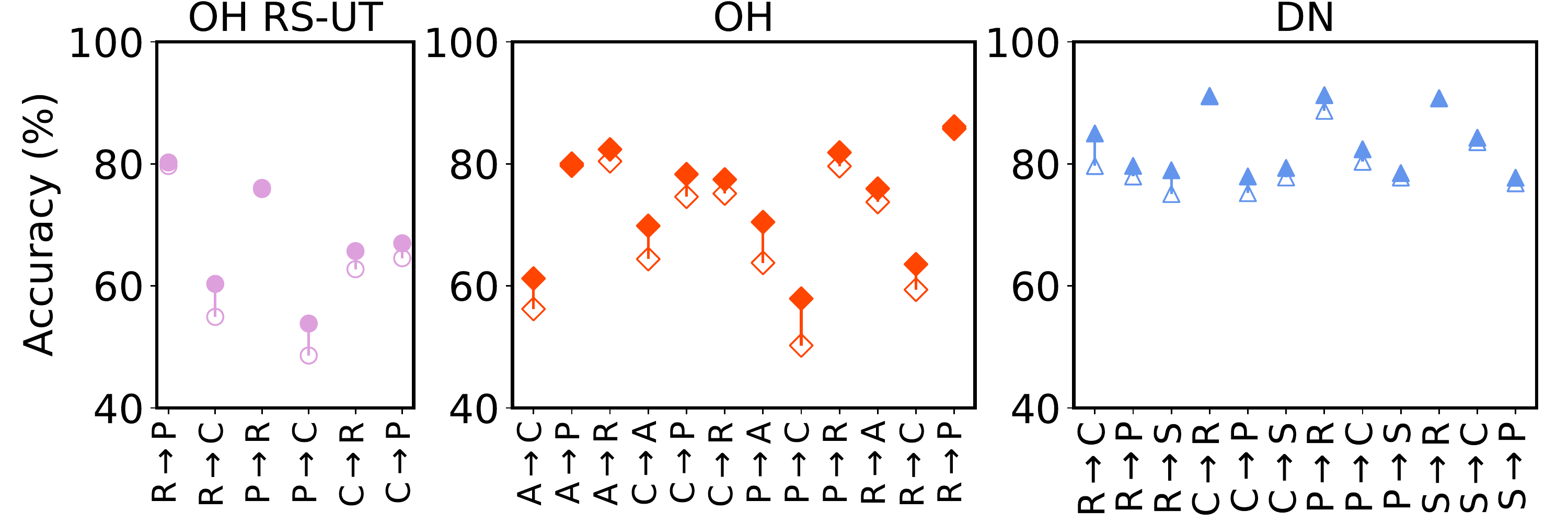}
	\end{center}
	\caption{Comparison between using activation normalization (solid markers) and not (hollow markers).}		
	\label{fig:norm}
\end{figure}

\textbf{Effects of activation normalization.}
As shown in Fig.~\ref{fig:norm}, using activation normalization consistently improves on all tasks. The performance gain is most significant when the domain gap is large (\eg, tasks with low accuracy). This validates that activation normalization can bring source and target distribution closer, thus reduce the impact of accumulated error in self-training.

\noindent\textbf{Comparison with other self-training losses.} To fairly compare with other self-training methods, we implement them under the same framework and hyper-parameters as APA. The only difference is the loss used for target samples. Results are listed in Tab.~\ref{tab:cmploss}. Among them, FixMatch and SENTRY require additional random augmented target samples. Our method achieves the best scores on all datasets.

\begin{table}[h]
	\footnotesize
	\centering
	\caption{Comparison with other self-training losses.}
	\scalebox{0.85}{
		\begin{tabular}{p{2cm}p{1.5cm}<{\centering}p{1.5cm}<{\centering}p{1.5cm}<{\centering}}
			\toprule
			loss &  DN &  OH RS-UT & OH \\ 
			\midrule 		
			ENT & 79.00 & 62.76 & 70.68\\
			MI & 78.38 & 64.55 & 69.81\\
			FixMatch & 76.35 & 62.12 & 69.46 \\
			SENTRY & 80.31 & 66.64 & 73.04 \\
			VAT & 74.76 & 60.37 & 66.05 \\
			\midrule
			APA$^{u}$ & 83.08 & \HL{67.19} & 73.76 \\
			APA$^{n}$ & \HL{83.88} & 66.87 & \HL{74.98} \\
			\bottomrule
	\end{tabular} }
	\label{tab:cmploss}
\end{table}


\noindent\textbf{Why projecting $\r^{(p_n)}$ back to unit sphere?} In Eq.~\ref{eq:rfn_proj}, we post-process $\r^{(p_n)}$ to ensure $\overline{f(\bm{x})}+\r^{(p_n)}$ of unit norm. An alternative way is to add another normalization operation without using perturbation projection by optimizing 
\begin{equation}\label{eq:loss_pn_var}
	\ell^{(p_n^\prime)}(\bm{x})=D\big[  g_\sigma (\overline{f(\bm{x})}),  g_\sigma (\overline{\overline{f(\bm{x})}+\r^{(p_n)}}) \big] 
\end{equation}
However, this will encounter similar shrinking effect of normalization as discussed. Table~\ref{tab:domainnet_proj} shows that using perturbation projection consistently performs better than this strategy on all DomainNet tasks.

\noindent\textbf{Effects of freezing classifier $g$.}
In previous experiments, we allow parameters of the classifier $g$ to update during the adaptation stage. Whereas $g$ and the corresponding decision boundaries in the feature space usually change negligibly. Some previous work \cite{liang2020we} freeze $g$ during adaptation stage. Table~\ref{tab:freeze} shows that the two strategies perform comparably well in our method.

\begin{table}[h]
	\footnotesize
	\centering
	\caption{Effects of freezing parameters of $g$ during the adaptation stage in APA$^{u}$.}
	\scalebox{0.85}{
		\begin{tabular}{p{1cm}p{2cm}<{\centering}p{2cm}<{\centering}p{2cm}<{\centering}}
			\toprule
			freeze & DN & OH RS-UT & OH  \\ 
			\midrule 		
			$\checkmark$ & \HL{83.32} & 66.34 &	\HL{73.80} \\
			$\times$ & 83.08 &	\HL{67.19} &	73.76 \\
			\bottomrule
	\end{tabular} }
	\label{tab:freeze}
\end{table}

\begin{figure}[h]
	\centering
	\includegraphics[width=0.98\linewidth]{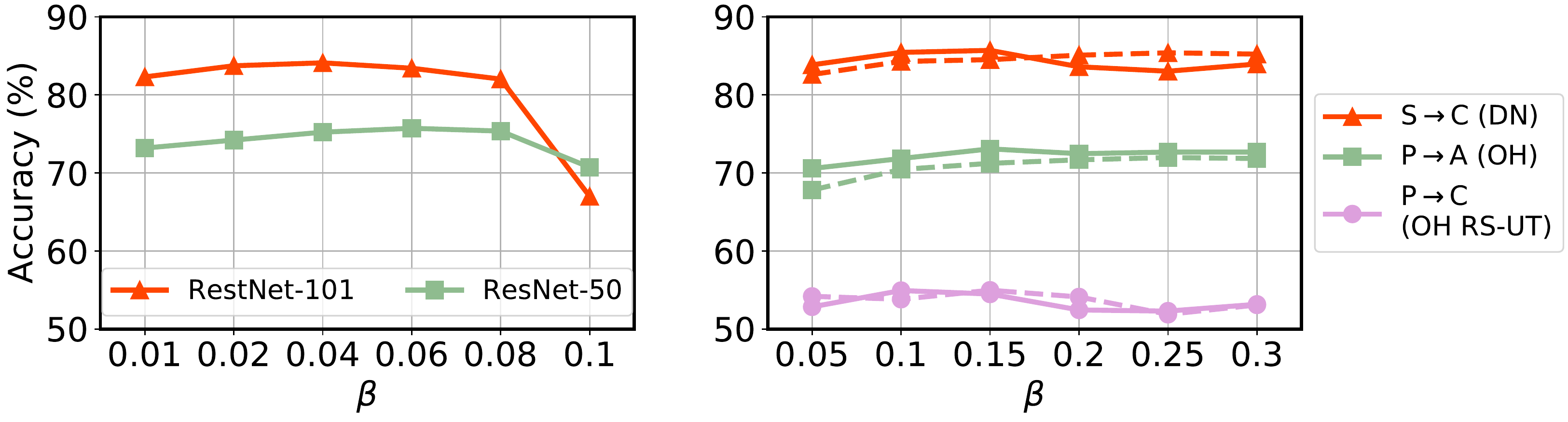}
	\caption{Effects of $\beta$ in APA$^{n}$ (solid lines) and APA$^{u}$ (dashed lines).}
	\label{fig:beta}
\end{figure}

\noindent\textbf{Sensitivity of hyper-parameters.} APA mainly involves two hyper-parameters, the perturbation magnitude $\epsilon$ and the adversarial loss weight $\beta$. Note that the average norm of penultimate activations is about 30. In Fig.~\ref{fig:eps_cmp}, We explicitly present results with large perturbations to show its robustness. Figure~\ref{fig:beta} plots the effects of $\beta$. Our method is insensitive to the $\epsilon$ and $\beta$ within a wide range.

\section{Conclusion}
This paper explores adversarial training on penultimate activations in UDA. We show its advantage through comparison with adversarial training on input images and intermediate features. Two variants are derived with activation normalization. Extensive experiments on popular benchmarks are conducted to show the superiority of our method over previous arts under both standard and source-data free setting. Our work demonstrates that adversarial training is a strong strategy in UDA tasks.

\section{Additional Implementation Details}
We use PyTorch 1.8.0 for implementation. Experiments are conducted on TITAN RTX. The feature extractor $f$ consists of a pretrained ResNet-50 or ResNet-101~\cite{he2016deep} backbone and a bottleneck module (\emph{Linear} $\rightarrow$ \emph{BatchNorm1d} $\rightarrow$ \emph{ReLU}). The classification head $g$ is a single \emph{Linear} layer. Following previous work~\cite{prabhu2021sentry}, we use a temperature of 0.05 for \texttt{softmax} activation. We first train the network on source data only with standard cross entropy loss, and then adapt it to the target data with the proposed objectives. In both stages, we use mini-batch SGD with momentum 0.8 and weight decay $5e^{-4}$. The learning rate $\eta$ is fixed to $1e^{-3}$ for the first stage, while in the second stage initialized as $\eta_{0}=1e^{-3}$ and scheduled by $\eta=\eta_{0}\cdot (1+1e^{-4}i)^{-0.75}$, where $i$ is the training step. Batch size is 32. For (pseudo) class-balanced sampling, we follow \cite{jiang2020implicit} to update target pseudo labels at an interval of 100 or 500 iterations. The computation cost could be further reduced by updating confident target samples less frequently. It is worth mentioning that $\xi$ is chosen as a small value of 1e-6 in VAT~\cite{miyato2018virtual} when approximating the optimal perturbation. However, we empirically find that using larger $\xi$ leads to slightly better performance. In our proposed method, we use $\xi=10$, $\epsilon=30$ for APA$^u$ and $\xi=1.0$, $\epsilon=1.0$ for APA$^n$.

\setlength{\tabcolsep}{3pt}
\begin{table*}[!t]
	\footnotesize
	\caption{Per-class average accuracies (\%) on \textbf{Domainnet}  (ResNet-50).}	
	\centering
	\scalebox{0.85}{
		\begin{tabular}{p{2cm}p{1.0cm}<{\centering}p{1.0cm}<{\centering}p{1.0cm}<{\centering}p{1.0cm}<{\centering}p{1.0cm}<{\centering}p{1.0cm}<{\centering}p{1.0cm}<{\centering}p{1.0cm}<{\centering}p{1.0cm}<{\centering}p{1.0cm}<{\centering}p{1.0cm}<{\centering}p{1.0cm}<{\centering}>{\columncolor{tbgray}}p{1.0cm}<{\centering}}
			\toprule
			Loss & R$\shortrightarrow$C &	R$\shortrightarrow$P &	R$\shortrightarrow$S &	C$\shortrightarrow$R &	C$\shortrightarrow$P &	C$\shortrightarrow$S &	P$\shortrightarrow$R &	P$\shortrightarrow$C &	P$\shortrightarrow$S &	S$\shortrightarrow$R &	S$\shortrightarrow$C &	S$\shortrightarrow$P	 & Avg.   \\ 	
			\midrule
			$\mathcal{L}_{\rm ENT}$ & 79.5 &	75.7 &	73.7 &	89.8 &	74.3 &	75.1 &	89.4 &	76.1 &	75.2 &	89.5 &	78.6 &	71.1 &	79.0 \\
			$\mathcal{L}_{\rm MI}$ & 78.3 &	74.7 &	72.6 &	88.7 &	73.5 &	75.2 &	88.5 &	75.7 &	74.0 &	87.6 &	77.2 &	74.6 &	78.4 \\
			$\mathcal{L}_{\rm 	FixMatch}$ & 75.1 &	76.7 &	68.0 &	88.9 &	72.4 &	73.3 &	89.2 &	73.2 &	70.2 &	85.3 &	75.1 &	68.8 &	76.4 \\
			$\mathcal{L}_{\rm SENTRY}$ & 80.2 &	77.5 &	74.4 &	90.2 &	73.6 &	78.5 &	88.0 &	78.8 &	77.9 &	89.9 &	82.3 &	72.4 &	80.3 \\
			$\mathcal{L}_{\rm VAT}$ & 73.6 &	74.8 &	65.2 &	88.1 &	66.1 &	64.5 &	88.2 &	71.4 &	70.7 &	87.3 &	77.4 &	69.9 &	74.8 \\		
			$\mathcal{L}_{\rm APA^u}$ & 85.0 &	79.7 &	79.0 &	91.2 &	\HL{77.9} &	79.3 &	91.3 &	82.4 &	78.5 &	90.7 &	84.3 &	77.7 &	83.1 \\
			$\mathcal{L}_{\rm APA^n}$ & \HL{85.3} &	\HL{80.4} &	\HL{80.2} &	\HL{91.8} &	77.2 &	\HL{79.9} &	\HL{91.7} &	\HL{83.7} &	\HL{80.4} &	\HL{91.2} &	\HL{85.4} &	\HL{79.3} &	\HL{83.9} \\
			\bottomrule
	\end{tabular} }
	\label{tab:sup:domainnet}
\end{table*}
\setlength{\tabcolsep}{1.4pt}

\setlength{\tabcolsep}{3pt}
\begin{table*}[!t]
	\footnotesize
	\caption{Accuracies (\%) on \textbf{Office-Home}  (ResNet-50).}	
	\centering
	\scalebox{0.85}{
		\begin{tabular}{p{2cm}p{1.0cm}<{\centering}p{1.0cm}<{\centering}p{1.0cm}<{\centering}p{1.0cm}<{\centering}p{1.0cm}<{\centering}p{1.0cm}<{\centering}p{1.0cm}<{\centering}p{1.0cm}<{\centering}p{1.0cm}<{\centering}p{1.0cm}<{\centering}p{1.0cm}<{\centering}p{1.0cm}<{\centering}>{\columncolor{tbgray}}p{1.0cm}<{\centering}}
			\toprule
			Loss & A$\shortrightarrow$C & A$\shortrightarrow$P & A$\shortrightarrow$R & C$\shortrightarrow$A & C$\shortrightarrow$P & C$\shortrightarrow$R &  P$\shortrightarrow$A & P$\shortrightarrow$C & P$\shortrightarrow$R & R$\shortrightarrow$A & R$\shortrightarrow$C & R$\shortrightarrow$P & Avg.   \\ 	
			\midrule
			$\mathcal{L}_{\rm ENT}$ & 56.6 &	78.3 &	81.3 &	66.3 &	77.3 &	76.0 &	65.8 &	52.6 &	81.1 &	71.4 &	57.4 &	84.2 &	70.7\\		
			$\mathcal{L}_{\rm MI}$ & 55.9 &	76.7 &	79.8 &	65.3 &	73.9 &	74.6 &	65.2 &	53.6 &	79.2 &	72.4 &	58.4 &	82.6 &	69.8 \\
			$\mathcal{L}_{\rm FixMatch}$ & 56.4 &	77.9 &	81.4 &	64.1 &	74.0 &	73.4 &	65.2 &	53.8 &	80.0 &	72.1 &	50.9 &	84.3 & 	69.5 \\
			$\mathcal{L}_{\rm SENTRY}$ & 60.7 &	79.9 &	\HL{82.6} &	66.7 &	78.0 &	\HL{79.6} &	66.7 &	57.3 &	82.6 &	73.8 &	63.4 &	85.1 &	73.0 \\
			$\mathcal{L}_{\rm VAT}$ & 	49.1 &	75.0 &	78.6 &	58.4 &	71.4 &	72.4 &	57.0 &	46.4 &	78.2 &	69.4 &	54.0 &	82.8 &	66.1 \\		
			$\mathcal{L}_{\rm APA^u}$ & 61.2 &	80.0 &	82.4 &	69.8 &	78.3 &	77.4 &	70.5 &	57.9 &	81.9 &	75.9 &	63.6 &	\HL{86.1} &	73.8 \\
			$\mathcal{L}_{\rm APA^n}$ & \HL{62.0} &	\HL{81.2} &	\HL{82.6} &	\HL{71.5} &	\HL{80.6} &	79.3 &	\HL{71.9} &	\HL{60.1} &	\HL{83.4} &	\HL{76.9} &	\HL{64.2} &	\HL{86.1} &	\HL{75.0} \\
			\bottomrule
	\end{tabular} }
	\label{tab:sup:officehome}
\end{table*}
\setlength{\tabcolsep}{1.4pt}

\begin{table*}[t]
	\footnotesize
	\centering
	\caption{Per-class average accuracies (\%) on \textbf{Office-Home (RS-UT)}  (ResNet-50).}
	\scalebox{0.85}{
		\begin{tabular}{@{}p{2.5cm}p{1.2cm}<{\centering}p{1.2cm}<{\centering}p{1.2cm}<{\centering}p{1.2cm}<{\centering}p{1.2cm}<{\centering}p{1.2cm}<{\centering}>{\columncolor{tbgray}}p{1.2cm}<{\centering}}
			\toprule
			Loss & R$\shortrightarrow$P &	R$\shortrightarrow$C &	P$\shortrightarrow$R &	P$\shortrightarrow$C &	C$\shortrightarrow$R & C$\shortrightarrow$P & Avg.   \\ 
			\midrule
			$\mathcal{L}_{\rm ENT}$ & 79.6 &	50.3 &	74.8 &	46.2 &	62.6 &	63.0 &	62.8 \\		
			$\mathcal{L}_{\rm MI}$ & 79.0 &	53.8 &	73.9 &	50.8 &	64.8 &	65.0 &	64.5 \\	
			$\mathcal{L}_{\rm FixMatch}$ & 77.2 &	51.5 &	73.8 &	47.8 &	59.9 &	62.5 &	62.1 \\
			$\mathcal{L}_{\rm SENTRY}$ & \HL{81.0} &	56.6 &	74.6 &	\HL{55.0} &	65.4 &	67.2 &	66.6 \\
			$\mathcal{L}_{\rm VAT}$ & 78.0 &	45.4 &	73.0 &	43.2 &	60.6 &	62.1 &	60.4 \\		
			$\mathcal{L}_{\rm APA^u}$ & 80.2 &	\HL{60.3} &	76.1 &	53.8 &	\HL{65.7} &	67.0 &	\HL{67.2} \\
			$\mathcal{L}_{\rm APA^n}$ & 80.3 &	57.7 &	\HL{76.7} &	54.9 &	64.1 &	\HL{67.5} &	66.9 \\		
			\bottomrule
	\end{tabular} }
	\label{tab:supp:officehome-rsut}
\end{table*}

\section{Proofs for \textit{Shrinking Effect of Normalization}}
APA$^u$ and APA$^n$ apply perturbation to the un-normalized and normalized activations, respectively. Their objectives are
\begin{equation}\label{eq:loss_fu}
	\ell^{(p_u)}(\bm{x})=D\big[  g_\sigma (\overline{f(\bm{x})}),  g_\sigma (\overline{f(\bm{x})+{\r^{(p_u)}}}) \big]
\end{equation}
\begin{equation}\label{eq:loss_fn}
	\ell^{(p_n)}(\bm{x})= D\big[  g_\sigma (\overline{f(\bm{x})}),  g_\sigma (\overline{f(\bm{x})}+{\r^{(p_n)}}) \big]
\end{equation}
Given the optimal perturbation $\r^{(p_u)}$ of APA$^u$, it is easy to map it into the perturbation on the normalized activations. The same is with $\r^{(p_n)}$ of APA$^n$. Let 
\begin{eqnarray}
	\label{eq:r_s2p}
	\r^{(p_{n\rightarrow u})} & \triangleq & \r^{(p_n)}\cdot \| f(\bm{x}) \| \\
	\label{eq:r_p2s}
	\r^{(p_{u\rightarrow n})} & \triangleq & \overline{f(\bm{x})+\r^{(p_u)}}-\overline{f(\bm{x})}
\end{eqnarray}
Then by plugging them into Eq.~\ref{eq:loss_fu} and Eq.~\ref{eq:loss_fn}, we have
\begin{equation}\label{eq:proop_n2u}
	\begin{aligned}
		&\ell^{(p_u)}(\bm{x})|_{\r^{(p_u)}=\r^{(p_{n\rightarrow u})}} \\
		&=D\big[  g_\sigma (\overline{f(\bm{x})}),  g_\sigma (\overline{f(\bm{x})+\r^{(p_{n\rightarrow u})}}) \big] \\
		&=D\big[  g_\sigma (\overline{f(\bm{x})}),  g_\sigma (\overline{(\overline{f(\bm{x})}+\r^{(p_n)})\cdot \|f(\bm{x}) \|  } ) \big] \\
		&=D\big[  g_\sigma (\overline{f(\bm{x})}),  g_\sigma (\overline{f(\bm{x})}+\r^{(p_n)}) \big] \\
		&=\ell^{(p_n)}(\bm{x})
	\end{aligned}
\end{equation}
\begin{equation}\label{eq:proop_u2n}
	\begin{aligned}
		&\ell^{(p_n)}(\bm{x})|_{\r^{(p_n)}=\r^{(p_{u\rightarrow n})}}\\
		&=D\big[  g_\sigma (\overline{f(\bm{x})}),  g_\sigma (\overline{f(\bm{x})}+\r^{(p_{u\rightarrow n})}) \big] \\
		&=D\big[  g_\sigma (\overline{f(\bm{x})}),  g_\sigma (\overline{f(\bm{x})}+\overline{f(\bm{x})+\r^{(p_u)}}-\overline{f(\bm{x})}) \big] \\
		&=D\big[  g_\sigma (\overline{f(\bm{x})}),  g_\sigma (\overline{f(\bm{x})+\r^{(p_u)}}) \big] \\
		&=\ell^{(p_u)}(\bm{x})
	\end{aligned}
\end{equation}
The third equality in Eq.~\ref{eq:proop_n2u} holds because $\overline{f(\bm{x})}+\r^{(p_n)}$ has unit norm due to perturbation projection in Eq. 9 of the paper. From Eq.~\ref{eq:proop_n2u} and Eq.~\ref{eq:proop_u2n}, $\ell^{(p_u)}|_{\r^{(p_u)}=\r^{(p_{n\rightarrow  u})}}$ equals $\ell^{(p_n)}$ and $\ell^{(p_n)}|_{\r^{(p_n)}=\r^{(p_{u\rightarrow n})}}$ equals $\ell^{(p_u)}$ in terms of objective value. However, their performance could be different due to different backpropagation. 

Let $\zeta(\bm{v})=\bm{v}/\|\bm{v}\|_2$ be the $\ell_2$ normalization function, $\x$ omitted, $\zeta_u\triangleq \zeta(f+\r^{(p_u)})$, and $\zeta_n\triangleq \zeta(f)+\r^{(p_n)}$. Based on the rules of gradients, we have  
\begin{equation}\label{eq:gf_u}
	\begin{aligned}
	\left(\frac{\partial \ell^{(p_u)}}{\partial f}\right)^{\top} &=  \left(\frac{\partial \ell^{(p_u)}}{\partial \zeta_u} \right)^{\top} \bm{J}_{\zeta_u}(f) \\
	 &= \left(\frac{\partial \ell^{(p_u)}}{\partial \zeta_u} \right)^{\top} \bm{J}_{\zeta}(f+\r^{(p_u)}) 
	\end{aligned}
\end{equation}
\begin{equation}\label{eq:gf_n}
    \begin{aligned}
	\left(\frac{\partial \ell^{(p_n)}}{\partial f}\right)^{\top} &=  \left(\frac{\partial \ell^{(p_n)}}{\partial \zeta_n} \right)^{\top} \bm{J}_{\zeta_n}(f) \\ 
	& = \left(\frac{\partial \ell^{(p_n)}}{\partial \zeta_n} \right)^{\top} \bm{J}_{\zeta}(f)
	\end{aligned}
\end{equation}

Since $\zeta_u|_{\r^{(p_u)}=\r^{(p_{n\rightarrow  u})}}=\zeta_n$ and $\zeta_n|_{\r^{(p_n)}=\r^{(p_{u\rightarrow  n})}}=\zeta_u$, it is easy to see that $\frac{\partial \ell^{(p_u)}}{\partial \zeta_u}|_{\r^{(p_u)}=\r^{(p_{n\rightarrow  u})}}=\frac{\partial \ell^{(p_n)}}{\partial \zeta_n}$ because the value of \textit{sources}  ($\zeta_u|_{\r^{(p_u)}=\r^{(p_{n\rightarrow  u})}}$ and $\zeta_n$), \textit{targets} ($\ell^{(p_u)}|_{\r^{(p_u)}=\r^{(p_{n\rightarrow u})}}$ and $\ell^{(p_n)}$), and \textit{mapping function} ($D\big[  g_\sigma (\overline{f(\bm{x})}),  g_\sigma (\cdot) \big]$) are all identical. Similarly, $\frac{\partial \ell^{(p_n)}}{\partial \zeta_n}|_{\r^{(p_n)}=\r^{(p_{u\rightarrow  n})}}=\frac{\partial \ell^{(p_u)}}{\partial \zeta_u}$. From Eq.~\ref{eq:gf_u}, we have
\begin{equation}\label{eq:derivative_u}
	\left(\frac{\partial \ell^{(p_u)}|_{\r^{(p_u)}=\r^{(p_{n\rightarrow  u})}}}{\partial f}\right)^{\top}=\left(\frac{\partial \ell^{(p_n)}}{\partial \zeta_n} \right)^{\top} \bm{J}_{\zeta}(f+\r^{(p_u)}) 
\end{equation}
Comparing Eq.~\ref{eq:derivative_u} with Eq.~\ref{eq:gf_n}, $\bm{J}_{\zeta}(f)$ in the right hand of Eq.~\ref{eq:gf_n} becomes $\bm{J}_{\zeta}(f+\r^{(p_u)})$ here. This results in different gradients of the two losses, despite that the perturbation and loss value are shared. Similarly from Eq.~\ref{eq:gf_n},
\begin{equation}\label{eq:derivative_n}
	\left(\frac{\partial \ell^{(p_n)}|_{\r^{(p_n)}=\r^{(p_{u\rightarrow  n})}}}{\partial f}\right)^{\top}=\left(\frac{\partial \ell^{(p_u)}}{\partial \zeta_u} \right)^{\top} \bm{J}_{\zeta}(f) 
\end{equation}
The right hand of Eq.~\ref{eq:derivative_n} is also different from Eq.\ref{eq:gf_u}.  

The above analyses demonstrate that due to reversed order of normalization and applying perturbation, APA$^u$ and APA$^n$ are related yet with different properties. The difference mainly comes from the non-linear $\ell_2$ normalization layer, which changes the gradients. It further leads to the shrinking effect in APA$^u$ as discussed in the paper.

\section{Comparison with Other Self-training Losses}
To fairly compare with other self-training methods, we implement several self-training losses under the same framework and hyper-parameters as APA. The objectives are all formulated as $\mathbb{E}_{(\bm{x}_s, y_s)}\ell_{\mathrm{ce}}(\bm{x}_s, y_s) + 0.1\mathcal{L}_{\mathrm{tgt}}$, where $\mathcal{L}_{\mathrm{tgt}}$ is instantiated with the corresponding losses. 
\begin{itemize}
	\item \textbf{ENT}: conditional entropy loss is widely used in semi-supervised learning to learn compact clusters. Let $\mathcal{H}(\x_t)=\sum_c -h_{\sigma_c}(\x_t)\log (h_{\sigma_c}(\x_t))$ denotes the conditional entropy of $\x_t$. The formulation of conditional entropy loss is 
	\begin{equation}
		\mathcal{L}_{\rm ENT}=\mathbb{E}_{\bm{x}_t} \mathcal{H}(\x_t)
	\end{equation}
	
	\item \textbf{MI}: conditional entropy loss may lead to trivial solution by classifying all samples to one category. Mutual Information Maximization loss~\cite{shi2012information} overcomes this drawback by introducing a diversity-promoting term. Its formulation is
	\begin{equation}
		\mathcal{L}_{\rm MI}=\mathcal{L}_{\rm ENT}+\sum_{c} \hat{p}_c \log (\hat{p}_c)
	\end{equation}
	where $\hat{p}=\mathbb{E}_{\bm{x}_t} h_{\sigma}(\x_t) $ is the mean model predicted probability.
	
	\item \textbf{VAT}: VAT~\cite{miyato2018virtual} adds perturbation to the input images. The loss formulation is 
	\begin{equation}
		\mathcal{L}_{\rm VAT}=\mathbb{E}_{\bm{x}_t}  D\big[ g_\sigma(\x_t), g_\sigma(\x_t+\r^{(v)}) \big]
	\end{equation}
	where $\r^{(v)}$ is the adversarial perturbation.
	
	\item \textbf{FixMatch}: FixMatch~\cite{sohn2020fixmatch} is a consistency regularization based method. It uses a weakly-augmented version $\alpha(\x_t)$ to supervise a strongly-augmented version $\mathcal{A}(\x_t)$. Let $q_b=h_{\sigma}(\alpha(\x_t))$ be the model predicted probability and $\hat{q}_b=\arg\max(q_b)$. The formulation of FixMatch loss is
	\begin{equation}
		\mathcal{L}_{\rm FixMatch}=-\mathbb{E}_{\bm{x}_t}  \mathbb{I}(\max(q_b)\geq \tau) \log (h_{\sigma_{\hat{q}_b}}(\mathcal{A}(\x_t))) )
	\end{equation}
	where $\tau$ is a confidence threshold.
	\item \textbf{SENTRY}: SENTRY~\cite{prabhu2021sentry} uses the predictive consistency under a committee of random image transformations of a target sample to judges its reliability. Then it selectively minimizes entropy of highly consistent target samples and maximizes entropy of inconsistent ones. It also adopts an ``information-entropy" loss to promote diversity. Let $\mathcal{D}^c$ denotes consistent samples and $\mathcal{D}^{i}$ denotes inconsistent ones. The formulation of SENTRY loss is
	\begin{equation}
	  \begin{aligned}
		\mathcal{L}_{\rm SENTRY}=&\mathbb{E}_{\bm{x}_t \in\mathcal{D}^c} \mathcal{H}(a_i(\x_t))-\mathbb{E}_{\bm{x}_t \in\mathcal{D}^i} \mathcal{H}(a_j(\x_t)) \\
		& + \mathbb{E}_{\x_t} \sum_c h_{\sigma_c}(\x_t)\log \tilde{q}_c
	\end{aligned}
	\end{equation}
	where $a_i(\cdot)$ and $a_j(\cdot)$ are random transformations, $\tilde{q}$ is the averaged predicted probability of the last $Q$ samples.
	
	\item \textbf{APA}: the loss formulations of APA$^u$ and APA$^n$ are 
	\begin{equation}
		\mathcal{L}_{\rm APA^u}=\mathbb{E}_{\bm{x}_t}  D\big[  g_\sigma (\overline{f(\bm{x_t})}),  g_\sigma (\overline{f(\bm{x_t})+{\r^{(p_u)}}}) \big]
	\end{equation}
	\begin{equation}
		\mathcal{L}_{\rm APA^n}=\mathbb{E}_{\bm{x}_t}  D\big[  g_\sigma (\overline{f(\bm{x}_t)}),  g_\sigma (\overline{f(\bm{x}_t)}+{\r^{(p_n)}}) \big] 
	\end{equation}
	where $\r^{(p_u)}$ and $\r^{(p_n)}$ are defined in Eq. 7 and Eq. 8 of the paper, respectively.
	
\end{itemize}

\noindent All these self-training losses aim to improve model confidence on unlabeled samples. Tables~\ref{tab:sup:domainnet}-\ref{tab:supp:officehome-rsut} list detailed results on three benchmarks that correspond to Tab. 6 of the paper. Note that SENTRY and FixMatch rely on additional random augmented images. Our method achieves consistently the best scores.

\section{Perturbation Using Top-k Labels}

Recall that in APA, the adversarial perturbation $\r^{(p_u)}$ and $\r^{(p_n)}$ are obtained by maximizing the divergence between model predictions of the original and perturbed activations. To understand their meaning, we rewrite their objectives as

\begin{equation}\label{eq:vatk_uo}
	\begin{aligned}
		\r^{(p_u)}&=\mathop{\arg\max}_{\|\r\|_2\leq \epsilon} ~ \ell^{(p_u)}_r(\r)=D\big[  g_\sigma (\overline{f(\bm{x})}),  g_\sigma (\overline{f(\bm{x})+\r}) \big] \\
		&=\mathop{\arg\max}_{\|\r\|_2\leq \epsilon}\ \sum_{c} - g_{\sigma_{c}}(\overline{f(\bm{x})})\ \log(  g_{\sigma_{c}}(\overline{f(\bm{x})+\r}) ) \\
		& \triangleq \mathop{\arg\max}_{\|\r\|_2\leq \epsilon}\ \sum_{c} w_c\ \ell_{r_c}^{(p_u)}(\r)
	\end{aligned}		
\end{equation}
and
\begin{equation}\label{eq:vatk_no}
	\begin{aligned}
		\r^{(p_n)}&=\mathop{\arg\max}_{\|\r\|_2\leq \epsilon} ~ \ell^{(p_n)}_r(\r)=D\big[  g_\sigma (\overline{f(\bm{x})}),  g_\sigma (\overline{\overline{f(\bm{x})}+\r}) \big] \\
		&=\mathop{\arg\max}_{\|\r\|_2\leq \epsilon}\ \sum_{c} - g_{\sigma_{c}}(\overline{f(\bm{x})})\ \log(  g_{\sigma_{c}}(\overline{\overline{f(\bm{x})}+\r}) ) \\
		& \triangleq \mathop{\arg\max}_{\|\r\|_2\leq \epsilon}\ \sum_{c} w_c\ \ell_{r_c}^{(p_n)}(\r)
	\end{aligned}		
\end{equation}
where $w_c \triangleq  g_{\sigma_{c}}(\overline{f(\bm{x})})$ means the $c$-th component of $g_{\sigma}(\overline{f(\x)})$, $\ell_{r_c}^{(p_u)}(\r)\triangleq-\log( g_{\sigma_{c}}(\overline{f(\bm{x})+\r}) )$, and $\ell_{r_c}^{(p_n)}(\r)\triangleq-\log( g_{\sigma_{c}}(\overline{\overline{f(\bm{x})}+\r}) )$. Both objectives in the \texttt{argmax} problems are weighted summations of several cross entropy losses.

\begin{figure}[!t]
	\centering
	\includegraphics[width=1\linewidth]{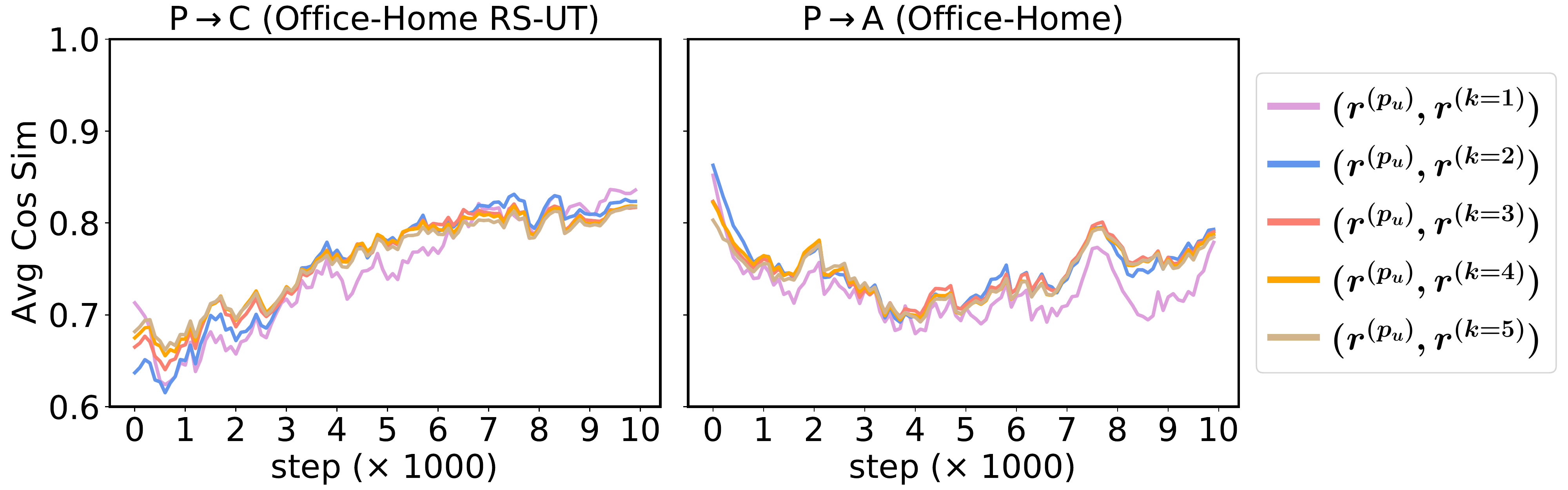}
    \includegraphics[width=1\linewidth]{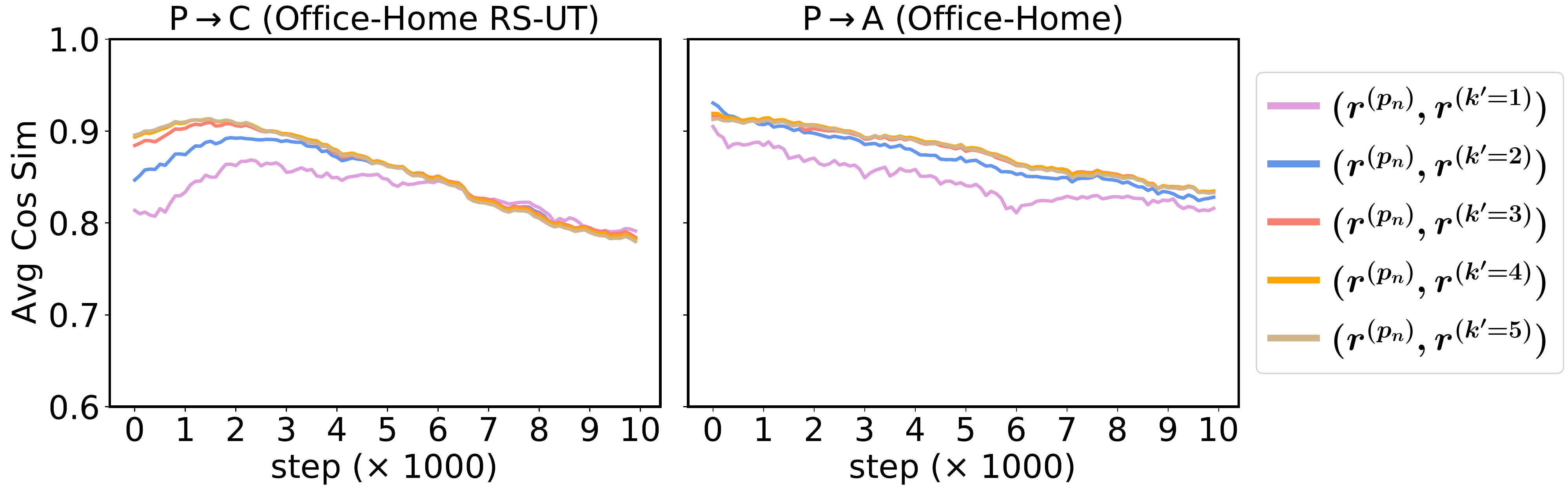}
	\caption{Averaged cosine similarity between $\r^{(p_u)}$ and $\r^{(k)}$ (upper), $\r^{(p_n)}$ and $\r^{(k^\prime)}$ (lower) during the training process of two UDA tasks. (Curves are smoothed for better visualization. See text for details).}
	\label{fig:supp:topk_corr}
\end{figure}

We truncate the summation by keeping only the terms corresponding to the top-k pseudo labels, and approximate $\r^{(p_u)}$ as
\begin{equation}\label{eq:vatk_u}
	\begin{split}
		\r^{(k)} &=\sum_{c=1}^{k} w_c \r^{(k_c)} \\
		\mathrm{ with~~}	\r^{(k_c)}&=\mathop{\arg\max}_{\|\r\|_2\leq \epsilon}\ \ell_{r_c}^{(p_u)}(\r)=-\log( g_{\sigma_c}(\overline{f(\bm{x})+\r}) )	
	\end{split}
\end{equation}
and approximate $\r^{(p_n)}$ as
\begin{equation}\label{eq:vatk_n}
	\begin{split}
		\r^{(k^\prime)} &=\sum_{c=1}^{k^\prime} w_c \r^{(k^\prime_c)} \\
		\mathrm{ with~~}	\r^{(k^\prime_c)}&=\mathop{\arg\max}_{\|\r\|_2\leq \epsilon}\ \ell_{r_c}^{(p_n)}(\r)=-\log( g_{\sigma_c}(\overline{\overline{f(\bm{x})}+\r}) )	
	\end{split}
\end{equation}
Eq.~\ref{eq:vatk_u}, and Eq.~\ref{eq:vatk_n} can be solved with projected gradient ascent from $\r=\bm{0}$. $\r^{(k)}$ and $\r^{(k^\prime)}$ has a clear meaning in pointing opposite to the top-k classes. The averaged cosine similarity between $\r^{(p_u)}$ and $\r^{(k)}$ are plotted in Fig.~\ref{fig:supp:topk_corr}. Overall, they have a strong positive correlation. The similarity is larger for bigger $k$. Similar trends are observed for $\r^{(p_n)}$ and $\r^{(k\prime)}$. As can be seen from Tab.~\ref{tab:supp:topk}, perturbation using top-k pseudo labels performs comparably to APA. These observations imply that it is beneficial to move samples closer to the top-k classes and away from the rest classes, as APA is motivated. 

\begin{table}[!t]
	\footnotesize
	\centering
	\caption{Adversarial feature perturbation based on top-k pseudo labels.}
	\scalebox{0.85}{
		\begin{tabular}{p{2cm}p{2cm}<{\centering}p{2cm}<{\centering}p{2cm}<{\centering}}
			\toprule
			& DN &  OH RS-UT & OH \\ 
			\midrule 	
			APA$^{k=1}$ & 82.83 &	67.18 &	73.72		 \\
			APA$^{k=2}$ & 83.35 &	67.49 &	74.38 \\
			APA$^{k=3}$ & 83.14 &	\HL{67.85} &	74.56	 \\
			APA$^{k=4}$ & 83.27 &	67.52 &	74.75 \\
			APA$^{k=5}$ & \HL{83.44} &	67.40 &	\HL{74.81} \\
			\midrule 
			APA$^u$ & 83.08 &	67.19 &	73.76 \\	
		
        \midrule
		\midrule 		
		APA$^{k^\prime=1}$ & 83.16 & 65.99	 & 74.04			 \\
		APA$^{k^\prime=2}$ & 83.63 & 66.10   & 74.78	 \\
		APA$^{k^\prime=3}$ & 83.46  & \HL{66.96}	 &	74.97	 \\
		APA$^{k^\prime=4}$ & 83.50 & 66.71	 &	74.87   \\
		APA$^{k^\prime=5}$ & 83.54 & 66.94	 & \HL{75.11}   \\
		\midrule 
		APA$^n$ & \HL{83.88} &	66.87 & 74.98	 \\
		\bottomrule
\end{tabular} }
	\label{tab:supp:topk}
\end{table}

\bibliography{paper}

\end{document}